\documentclass[superscriptaddress,twocolumn]{revtex4}

\usepackage{url}
\usepackage{graphicx,epsfig,verbatim,enumerate}
\usepackage{amssymb,amsmath}

\usepackage{morefloats}

\usepackage{listings}
\usepackage{color}

\definecolor{lightgrey}{rgb}{0.5,0.5,0.5}

\usepackage{changepage}
\newcommand{\pnastableadjust}{0}

\usepackage{rotating}
\usepackage{longtable}
\usepackage{array}

\newcommand{\sindex}[1]{}
\newcommand{\nindex}[1]{}

\newcommand{\etal}{\textit{et al.}}
\newcommand{\www}[1]{\url{#1}}

\newcommand{\PreserveBackslash}[1]{\let\temp=\\#1\let\\=\temp}
\newcommand{\PBS}[1]{\let\temp=\\#1\let\\=\temp}

\newcommand{\plainlatexonly}[1]{}
\newcommand{\revtexlatexswitch}[2]{#1}
\newcommand{\plosoneonly}[1]{}

\begin{document}

\title{
Benchmarking sentiment analysis methods for large-scale texts:
A case for using continuum-scored words and word shift graphs.
}

\author{Andrew J. Reagan}
\affiliation{Department of Mathematics \& Statistics, Computational Story Lab, \& the Vermont Advanced Computing Core, University of Vermont, Burlington, VT, 05405}
\affiliation{Vermont Complex Systems Center, University of Vermont, Burlington, VT, 05405}
\author{Christopher M. Danforth}
\affiliation{Department of Mathematics \& Statistics, Computational Story Lab, \& the Vermont Advanced Computing Core, University of Vermont, Burlington, VT, 05405}
\affiliation{Vermont Complex Systems Center, University of Vermont, Burlington, VT, 05405}
\author{Brian Tivnan}
\affiliation{Vermont Complex Systems Center, University of Vermont, Burlington, VT, 05405}
\affiliation{The MITRE Corporation, 7525 Colshire Drive, McLean, VA, 22102}
\author{Jake Ryland Williams}
\affiliation{School of Information, University of California, Berkeley, Berkeley, CA, 94720}
\author{Peter Sheridan Dodds}
\affiliation{Department of Mathematics \& Statistics, Computational Story Lab, \& the Vermont Advanced Computing Core, University of Vermont, Burlington, VT, 05405}
\affiliation{Vermont Complex Systems Center, University of Vermont, Burlington, VT, 05405}

\date{\today}

\begin{abstract}
The emergence and global adoption of social media 
has rendered possible
the real-time estimation of population-scale sentiment,
issuing profound implications for 
our understanding of human behavior.
Given the growing assortment of sentiment measuring instruments,
comparisons between them are evidently required.
Here, we perform detailed, quantitative tests and qualitative assessments of 6 dictionary-based methods 
applied to 4 different corpora, and briefly examine a further 20 methods.
We show that a dictionary-based method will only perform
both reliably and meaningfully if 
(1) 
the dictionary covers a sufficiently large enough
portion of a given text's lexicon when weighted by word usage frequency;
and 
(2) 
words are scored on a continuous scale.
\end{abstract}

\maketitle

\section{Introduction}

As we move further into what might be called the Sociotechnocene---with 
increasingly more interactions, decisions, and impact being made
by globally distributed people and algorithms---the myriad human 
social dynamics that have shaped our history have become
far more visible and measurable than ever before.
Driven by the broad implications of being able to characterize social
systems in microscopic detail,
sentiment detection for populations at
all scale has become
a prominent research arena.
Attempts to leverage online expression for
sentiment mining include 
prediction of stock markets ~\cite{bollen2011twitter,si2013exploiting,chung2011predicting,ruiz2012correlating}, assessing responses to advertising, real-time monitoring of global happiness~\cite{dodds2015a}, and measuring a health-related quality of life~\cite{alajajian2015a}.
The diverse set of instruments produced by this work now provide indicators
that help scientists understand collective behavior, 
inform public policy makers, 
and in industry, 
gauge the sentiment of public response to marketing campaigns.
Given their widespread usage and potential to influence social systems,
understanding how these instruments perform and how they compare with each other has become an imperative.
Benchmarking their performance both focuses future development and provides practical advice to non-experts in selecting a dictionary.

We identify sentiment detection methods as
belonging to one of three categories, each carrying their own advantages and
disadvantages: 
\begin{enumerate}
\item 
  Dictionary-based methods~\cite{dodds2015a,bradley1999a,pennebaker2001a,wilson2005a,liu2010a,warriner2014a},
\item 
  Supervised learning methods~\cite{liu2010a}, 
  and 
\item 
  Unsupervised (or deep) learning methods~\cite{socher2013a}.
\end{enumerate}

Here, we focus on dictionary-based methods,
which all center around
the determination of a text $T$'s average happiness (sometimes referred to as \textit{valence})
through the equation:
\begin{equation} 
  h_\textnormal{avg}(T) = 
  \frac{
    \sum^{N}_{i=1} 
    h_{\textnormal{avg}}(w_i)
    \cdot 
    f_i (T)
  }
  {
    \sum^{N}_{i=1} 
    f_i (T)
  }
  = 
  \sum^{N}_{i=1} 
  h_\textnormal{avg} (w_i)
  \cdot 
  p_i (T),
  \label{eq:havg} 
\end{equation}
where we denote each of the $N$ words in a given dictionary as $w_i$,
word sentiment scores as $h_\textnormal{avg}(w_i)$, 
word frequency as $f_i(T)$,
and normalized frequency of $w_i$ in $T$ as
$
p_i (T) 
=
f_i (T) 
/ 
\sum^{N}_{i=1} 
f_i (T)
$.
In this way, we measure the happiness of a text
in a manner analogous to taking the temperature of a room.
While other simple happiness scores may be considered, we will see that
analyzing individual word contributions is important and that this
equation allows for a straightforward, meaningful interpretation.

Dictionary-based methods rely upon two
distinct advantages we will capitalize on: (1) they are in principle
corpus agnostic (including those without training data available) and 
(2) in contrast to black box (highly non-linear) methods, they offer
the ability to ``look under the hood'' at words contributing to a
particular score through ``word shifts'' (defined fully later; see
also ~\cite{dodds2009b,dodds2011a}).
Indeed, if we are at all concerned with understanding why a particular
scoring method 
varies---e.g,, our undertaking is scientific---then word shifts are
essential tools.
In the absence of word shifts or similar, any explanation of sentiment trends
is missing crucial information and rises only to the level of opinion
or guesswork~\cite{golder2011a,garcia2015a,dodds2015b,wojcik2015conservatives}.

As all methods must, dictionary-based ``bag-of-words'' approaches suffer
from various drawbacks, and three are worth stating up front.
First, they are only applicable to corpora of
sufficient size, well beyond that of a single sentence
(widespread usage in this misplaced fashion does not suffice as a counterargument).
We directly verify this assertion on individual Tweets, finding that some dictionaries perform admirably, however the average (median) F1-score on the STS-Gold data set is 0.50 (0.54) from all datasets (Table \ref{tbl:STS}), others having shown similar results for dictionary methods with short text \cite{ribeiro2016sentibench}.
Second, state-of-the-art learning methods with a sufficiently large training set for a specific corpus
will outperform dictionary-based methods on same corpus \cite{liu2012sentiment}.
However, in practice the domains and topics to which sentiment
analysis
are applied are highly varied, such that training to a high degree of
specificity for a single corpus may not be practical and, from
a scientific standpoint, will severely constrain attempts to detect
and understand universal patterns.
Third: words may be evaluated out of context or with the wrong meaning.
A simple example is the word ``new'' occurring frequently
when evaluating articles in the New York Times.
This kind of contextual error
is something we can readily identify and correct for through word shift graphs,
but would remain hidden to nonlinear learning methods without new training.

We lay out our paper as follows.
We list and describe the dictionary-based 
methods we consider in \revtexlatexswitch{Sec.~\ref{sec:dictionariesandcorpora}}{Sec.~Dictionaries,~Corpora,~and~Word~Shift~Graphs},
and outline the corpora we use for tests
in 
\revtexlatexswitch{Sec.~\ref{subsec:benchmarks}}{Subsec.~Corpora~Tested}.
We present our results in
\revtexlatexswitch{Sec.~\ref{sec:results}}{Sec.~Results},
comparing all methods in how they perform
for specific analyses of
the New York Times (NYT)
(\revtexlatexswitch{Sec.~\ref{subsec:NYTwordshift}}{Subsec.~New~{Y}ork~{T}imes~Word~Shift~Analysis}),
movie reviews
(\revtexlatexswitch{Sec.~\ref{subsec:moviereviews}}{Subsec.~Movie~{R}eviews~Classification~and~Word~Shift~Analysis}),
Google Books
(\revtexlatexswitch{Sec.~\ref{subsec:googlebooks}}{Subsec.~Google~{B}ooks~Time~Series~and~Word~Shift~Analysis}),
and 
Twitter
(\revtexlatexswitch{Sec.~\ref{subsec:twittertimeseries}}{Subsec.~Twitter~Time~Series~Analysis}).
In \revtexlatexswitch{Sec.~\ref{subsec:NB-section}}{Subsec.~Brief~Comparison~to~Machine~Learning~Methods}, we make
some basic comparisons between dictionary-based methods and machine learning approaches.
We bolster our findings with figures in the Supporting Information,
and provide concluding remarks in \revtexlatexswitch{Sec.~\ref{sec:conclusion}}{Sec.~Conclusion}.

\section{Dictionaries, Corpora, and Word Shift Graphs}
\label{sec:dictionariesandcorpora}

\begin{table*}[tbp!]
  \begin{adjustwidth}{\pnastableadjust in}{0in}
  {\scriptsize
  \begin{tabular*}{\linewidth}{ l | l | l | l | l | l | l | l | l | l}
    \hline
    Dictionary & \# Fixed & \# Stems & Total & Range & \# Pos & \# Neg & Construction & License & Ref.\\
    \hline
    \hline
    labMT & 10222 & 0 & 10222 & 1.3 $\to$ 8.5 & 7152 & 2977 & Survey: MT, 50 ratings & CC & \cite{dodds2015a}\\
    ANEW & 1034 & 0 & 1034 & 1.2 $\to$ 8.8 & 584 & 449 & Survey: FSU Psych 101 & Free for research & \cite{bradley1999a}\\
    LIWC07 & 2145 & 2338 & 4483 & [-1,0,1] & 406 & 500 & Manual & Paid, commercial & \cite{pennebaker2001a}\\
    MPQA & 5587 & 1605 & 7192 & [-1,0,1] & 2393 & 4342 & Manual + ML & GNU GPL & \cite{wilson2005a}\\
    OL & 6782 & 0 & 6782 & [-1,1] & 2003 & 4779 & Dictionary propagation & Free & \cite{liu2010a}\\
    WK & 13915 & 0 & 13915 & 1.3 $\to$ 8.5 & 7761 & 5945 & Survey: MT, at least 14 ratings & CC & \cite{warriner2014a}\\
\hline
    LIWC01 & 1232 & 1090 & 2322 & [-1,0,1] & 266 & 344 & Manual & Paid, commercial & \cite{pennebaker2001a}\\
    LIWC15 & 4071 & 2478 & 6549 & [-1,0,1] & 642 & 746 & Manual & Paid, commercial & \cite{pennebaker2001a}\\
    PANAS-X & 20 & 0 & 20 & [-1,1] & 10 & 10 & Manual & Copyrighted paper & \cite{watson1999panas}\\
    Pattern & 1528 & 0 & 1528 & -1.0 $\to$ 1.0 & 575 & 679 & Unspecified & BSD & \cite{de2012pattern}\\
    SentiWordNet & 147700 & 0 & 147700 & -1.0 $\to$ 1.0 & 17677 & 20410 & Synset synonyms & CC BY-SA 3.0 & \cite{baccianella2010sentiwordnet}\\
    AFINN & 2477 & 0 & 2477 & [-5,-4, $\ldots$,4,5] & 878 & 1598 & Manual & ODbL v1.0 & \cite{nielsen2011new}\\
    GI & 3629 & 0 & 3629 & [-1,1] & 1631 & 1998 & Harvard-IV-4 & Unspecified & \cite{stone1966general}\\
    WDAL & 8743 & 0 & 8743 & 0.0 $\to$ 3.0 & 6517 & 1778 & Survey: Columbia students & Unspecified & \cite{whissell1986dictionary}\\
    EmoLex & 14182 & 0 & 14182 & [-1,0,1] & 2231 & 3243 & Survey: MT & Free for research & \cite{mohammad2013crowdsourcing}\\
    MaxDiff & 1515 & 0 & 1515 & -1.0 $\to$ 1.0 & 775 & 726 & Survey: MT, MaxDiff & Free for research & \cite{kiritchenko2014sentiment}\\
    HashtagSent & 54129 & 0 & 54129 & -6.9 $\to$ 7.5 & 32048 & 22081 & PMI with hashtags & Free for research & \cite{zhu2014nrc}\\
    Sent140Lex & 62468 & 0 & 62468 & -5.0 $\to$ 5.0 & 38312 & 24156 & PMI with emoticons & Free for research & \cite{MohammadKZ2013}\\
    SOCAL & 7494 & 0 & 7494 & -30.2 $\to$ 30.7 & 3325 & 4169 & Manual & GNU GPL & \cite{taboada2011lexicon}\\
    SenticNet & 30000 & 0 & 30000 & -1.0 $\to$ 1.0 & 16715 & 13285 & Label propogation & Citation requested & \cite{cambria2014senticnet}\\
    Emoticons & 132 & 0 & 132 & [-1,0,1] & 58 & 48 & Manual & Open source code & \cite{gonccalves2013comparing}\\
    SentiStrength & 1270 & 1345 & 2615 & [-5,-4, $\ldots$,4,5] & 601 & 2002 & LIWC+GI & Unknown & \cite{thelwall2010sentiment}\\
    VADER & 7502 & 0 & 7502 & -3.9 $\to$ 3.4 & 3333 & 4169 & MT survey, 10 ratings & Freely available & \cite{hutto2014vader}\\
    Umigon & 927 & 0 & 927 & [-1,1] & 334 & 593 & Manual & Public Domain & \cite{levallois2013umigon}\\
    USent & 592 & 0 & 592 & [-1,1] & 63 & 529 & Manual & CC & \cite{pappas2013distinguishing}\\
    EmoSenticNet & 13188 & 0 & 13188 & [-10,-2,-1,0,1,10] & 9332 & 1480 & Bootstrapped extension & Non-commercial & \cite{poria2013enhanced}\\
  \end{tabular*}}
  \end{adjustwidth}
  \caption{
    Summary of dictionary attributes used in sentiment measurement instruments.
    We provide all acronyms and abbreviations and further information
    regarding dictionaries in \revtexlatexswitch{Sec.~\ref{subsec:dictionaries}}{Subsec.~Dictionaries}.
    We test the first 6 dictionaries extensively.
    \# Fixed, \# Stems, \# Pos and \# Neg refer
    to the numbers of: 
    terms in the dictionary that are
    fixed words,
    stems used to match words,
    terms that are rated above neutral, and terms rated below neutral.
    The range indicates whether scores are continuous or binary (we
    use the term binary for dictionaries for which words 
    are scored as $\pm 1$ and optionally 0).
  }
\label{tbl:summary}
\end{table*}

\subsection{Dictionaries}
\label{subsec:dictionaries}

The words ``dictionary,'' ``lexicon,'' and ``corpus'' are often used interchangeably, and for clarity we define our usage as follows.

\begin{description} \itemsep1pt \parskip1pt \parsep0pt
\item[Dictionary:] Set of words (possibly including word stems) with ratings.
\item[Corpus:] Collection of texts which we seek to analyze.
\item[Lexicon:] The words contained within a corpus (often said to be ``tokenized'').
\end{description}

We test the following six dictionaries in depth:

\begin{description} \itemsep1pt \parskip1pt \parsep0pt
\item[labMT] --- language assessment by Mechanical Turk~\cite{dodds2015a}.
\item[ANEW] --- Affective Norms of English Words~\cite{bradley1999a}.
\item[WK] --- Warriner and Kuperman rated words from SUBTLEX by Mechanical Turk~\cite{warriner2014a}.
\item[MPQA] --- The Multi-Perspective Question Answering (MPQA) Subjectivity Dictionary~\cite{wilson2005a}.
\item[LIWC\{01,07,15\}] --- Linguistic Inquiry and Word Count, three versions~\cite{pennebaker2001a}.
\item[OL] --- Opinion Lexicon, developed by Bing Liu~\cite{liu2010a}.
\end{description}

We also make note of 18 other dictionaries:
\begin{description} \itemsep1pt \parskip1pt \parsep0pt
  \item[PANAS-X] --- The Positive and Negative Affect Schedule --- Expanded \cite{watson1999panas}.
    \item[Pattern] --- A web mining module for the Python programming language, version 2.6 \cite{de2012pattern}.
    \item[SentiWordNet] --- WordNet synsets each assigned three sentiment scores: positivity, negativity, and objectivity \cite{baccianella2010sentiwordnet}.
    \item[AFINN] --- Words manually rated -5 to 5 with impact scores by Finn Nielsen \cite{nielsen2011new}.
    \item[GI] --- General Inquirer: database of words and manually created semantic and cognitive categories, including positive and negative connotations \cite{stone1966general}.
    \item[WDAL] --- Whissel's Dictionary of Affective Language: words rated in terms of their Pleasantness, Activation, and Imagery (concreteness) \cite{whissell1986dictionary}.
    \item[EmoLex] --- NRC Word-Emotion Association Lexicon: emotions and sentiment evoked by common words and phrases using Mechanical Turk \cite{mohammad2013crowdsourcing}.
    \item[MaxDiff] --- NRC MaxDiff Twitter Sentiment Lexicon: crowdsourced real-valued scores using the MaxDiff method \cite{kiritchenko2014sentiment}.
    \item[HashtagSent] --- NRC Hashtag Sentiment Lexicon: created from tweets using Pairwise Mutual Information with sentiment hashtags as positive and negative labels (here we use only the unigrams) \cite{zhu2014nrc}.
    \item[Sent140Lex] --- NRC Sentiment140 Lexicon: created from the ``sentiment140'' corpus of tweets, using Pairwise Mutual Information with emoticons as positive and negative labels (here we use only the unigrams) \cite{MohammadKZ2013}.
    \item[SOCAL] --- Manually constructed general-purpose sentiment dictionary \cite{taboada2011lexicon}.
    \item[SenticNet] --- Sentiment dataset labeled with semantics and 5 dimensions of emotions by Cambria \etal, version 3 \cite{cambria2014senticnet}.
    \item[Emoticons] --- Commonly used emoticons with their positive, negative, or neutral emotion \cite{gonccalves2013comparing}.
    \item[SentiStrength] --- an API and Java program for general purpose sentiment detection (here we use only the sentiment dictionary) \cite{thelwall2010sentiment}.
    \item[VADER] --- method developed specifically for Twitter and social media analysis \cite{hutto2014vader}.
    \item[Umigon] --- Manually built specifically to analyze Tweets from the sentiment140 corpus \cite{levallois2013umigon}.
    \item[USent] --- set of emoticons and bad words that extend MPQA \cite{pappas2013distinguishing}.
    \item[EmoSenticNet] --- extends SenticNet words with WNA labels \cite{poria2013enhanced}.
\end{description}

All of these dictionaries were produced by academic groups, and with the exception of LIWC, they are provided free of charge.
In Table~\ref{tbl:summary},
we supply the main aspects---such as word count, 
score type (continuum or binary), and license information---for 
the dictionaries listed above.
In the github repository associated with our paper,
\url{https://github.com/andyreagan/sentiment-analysis-comparison},
we include all of the dictionaries but LIWC.

The LabMT, ANEW, and WK dictionaries have scores ranging on a continuum from 1 (low happiness) to 9 (high happiness) with 5 as neutral, whereas the others we test in detail have scores of $\pm 1$, and either explicitly or implicitly 0 (neutral).
We will refer to the latter dictionaries as being binary, even if neutral is included.
Other non-binary ranges include a continuous scale from -1 to 1 (SentiWordNet), integers from -5 to 5 (AFINN), continuous from 1 to 3 (GI), and continuous from -5 to 5 (NRC).
For coverage tests, we include all available words, to gain a full sense of the breadth of each dictionary.
In scoring, we do not include neutral words from any dictionary.

We test the LabMT, ANEW, and WK dictionaries for a range of stop words (starting with the removal of words scoring within $\Delta_{h} = 1$ of the neutral score of 5)~\cite{dodds2011a}.
The ability to remove stop words is one advantage of dictionaries that have a range of scores, allowing us to tune the instrument for maximum performance, while retaining all of the benefits of a dictionary method.
We will show that, in agreement with the original paper introducing LabMT and looking at Twitter data, a $\Delta_{h} = 1$ is a pragmatic choice in general~\cite{dodds2011a}.

Since we do not apply a part of speech tagger, when using the MPQA
dictionary we are obliged to exclude words with scores of both +1 and -1.
The words and stems with both scores are: blood, boast* (we denote
stems with an asterisk), conscience, deep, destiny, keen, large, and precious.
We choose to match a text's words using
the fixed word set from each dictionary before stems, 
hence words with overlapping matches (a fixed word that also matches a
stem) are first matched by the fixed word.

\subsection{Corpora Tested}
\label{subsec:benchmarks}

For each dictionary, we test both the coverage and the ability to detect previously observed and/or known patterns within each of the following corpora, noting the pattern we hope to discern:
\begin{enumerate} 
  \itemsep1pt \parskip1pt \parsep0pt
\item 
  The New York Times (NYT)~\cite{nytimescorpus2008a}: Goal of ranking sections by sentiment (\revtexlatexswitch{Sec.~\ref{subsec:NYTwordshift}}{Subsec.~New~{Y}ork~{T}imes~Word~Shift~Analysis}).
\item 
  Movie reviews~\cite{pang2004a}: Goal of discerning positive and negative reviews (\revtexlatexswitch{Sec.~\ref{subsec:moviereviews}}{Subsec.~Movie~{R}eviews~Classification~and~Word~Shift~Analysis}).
\item 
  Google Books~\cite{lin2012syntactic}: Goal of creating time series (\revtexlatexswitch{Sec.~\ref{subsec:googlebooks}}{Subsec.~Google~{B}ooks~Time~Series~and~Word~Shift~Analysis}).
\item 
  Twitter: Goal of creating time series (\revtexlatexswitch{Sec.~\ref{subsec:twittertimeseries}}{Subsec.~Twitter~Time~Series~Analysis}).
\end{enumerate}

For the corpora other than the movie reviews and small numbers of tagged Tweets, there is no publicly available ground
truth sentiment, so we instead make comparisons between methods
and examine how words contribute to scores.
We note that comparison to societal measures of well being would also
be possible~\cite{Mitchell2013a}.
We offer greater detail on corpus processing below,
and we also provide the relevant scripts on github at
\url{https://github.com/andyreagan/sentiment-analysis-comparison}.

\subsection{Word Shift Graphs}
\label{subsec:wordshifts}

Sentiment analysis is applied in many circumstances in which the goal of the analysis goes beyond simply categorizing text into positive or negative.
Indeed if this were the only use case, the value added by sentiment analysis would be severely limited.
Instead we use sentiment analysis methods as a lens that allow us to see how the emotive words in a text shape the overall content.
This is accomplished by analyzing each word for the individual contribution to the sentiment score (or to the difference in the sentiment score between two texts).
In either case, we need to consider the words ranked by this individual contribution.

Of the four corpora that we use as benchmarks, three rely on this type of qualitative analysis: using the dictionary as a tool to better understand the sentiment of the corpora.
For this case, we must first find the contribution of each word individually.
Starting with two texts, we take the difference of their sentiment scores, rearrange a few things, and arrive at
\[
  h^{\textrm{(comp)}}_{\textrm{avg}} - h^{\textrm{(ref)}}_{\textrm{avg}}
  =
  \sum_{w \in L}
  \underbrace{
  \left[
  h_{\textrm{avg}} {(w)} - h^{\textrm{(ref)}}_{\textrm{avg}}
  \right]
  }_{+/-}
  \underbrace{
  \left[
  p_w^{\textrm{(comp)}} - p_w^{\textrm{(ref)}}
  \right]
    }_{\uparrow/\downarrow}.
  \]
  Each word contributes to the word shift according to its happiness relative
  to the reference text ($+/-$ = more/less emotive),
  and its change in frequency of usage ($\uparrow/\downarrow$ = more/less frequent).
  As a first step, it is possible to visualize this sorted word list in a table, along with the basic indicators of how its contribution is constituted.
  We use word shift graphs to present this information in the most accessible manner to advanced users.

  For further detail, we refer the reader to our instructional post and video at \href{http://www.uvm.edu/storylab/2014/10/06/hedonometer-2-0-measuring-happiness-and-using-word-shifts/}{http://www.uvm.edu/storylab/2014/10/06/}.
\section{Results}
\label{sec:results}

\begin{figure*}[tbp!]
  \centering
    \includegraphics[width=0.98\textwidth]{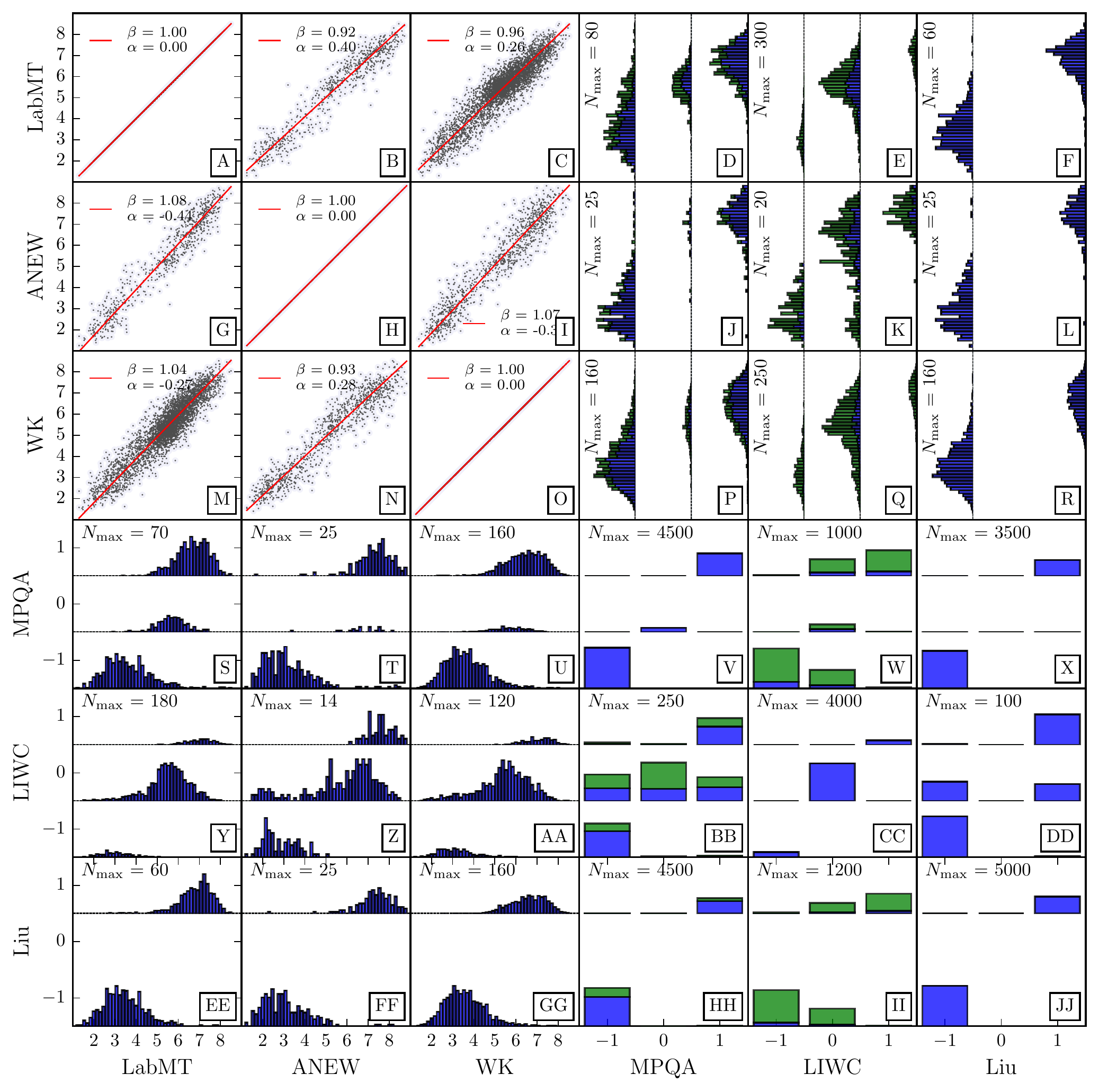}
  \caption{
    Direct comparison of the words in each of the dictionaries tested. For the comparison of two dictionaries, we plot words that are matched by the independent variable ``$x$'' in the dependent variable ``$y$''.
    Because of this, and cross stem matching, the plots are not symmetric across the diagonal of the entire figure.
    Where the scores are continuous in both dictionaries, we compute the RMA linear fit.
    When a dictionary contains both fixed and stem words, we plot the matches by fixed words in blue and by stem words in green.
    The axes in the bar plots are not of the same height, due to large mismatches in the number of words in the dictionaries, and we note the maximum height of the bar in the upper left of such plots.
    Detailed analysis of Panel C can be found in~\cite{dodds2015b}.
    We provide a table for each off-diagonal panel in the S2 Appendix with the words whose scores exhibit the greatest mismatch, and a subset of these tables in Fig~\ref{fig:labMT-tables}.
  }
  \label{fig:main}
\end{figure*}

\begin{figure*}[tbp!]
  \includegraphics[width=1.05\textwidth]{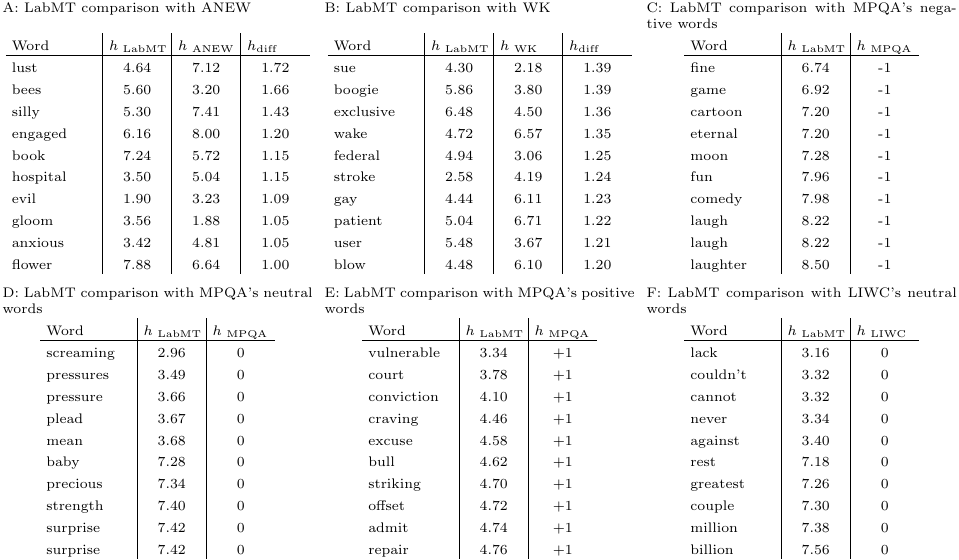}
    \caption{
      We present the specific words from Panels G, M, S and Y of Fig~\ref{fig:main} with the greatest mismatch.
      Only the center histogram from Panel Y of Fig~\ref{fig:main} is included.
      We emphasize that the LabMT dictionary scores generally agree well with the other dictionaries, and we are looking at the marginal words with the strongest disagreement.
      Within these words, we detect differences in the creation of these dictionaries that carry through to these edge cases.
      Panel A: The words with most different scores between the LabMT and ANEW dictionaries are suggestive of the different meanings that such words entail for the different demographic surveyed to score the words.
      Panel B: Both dictionaries use surveys from the same demographic (Mechanical Turk), where the LabMT dictionary required more individual ratings for each word (at least 50, compared to 14) and appears to have dampened the effect of multiple meaning words.
      Panels C--E: The words in LabMT matched by MPQA with scores of -1, 0, and +1 in MPQA show that there are at least a few words with negative rating in MPQA that are not negative (including the happiest word in the LabMT dictionary: ``laughter''), not all of the MPQA words with score 0 are neutral, and that MPQA's positive words are mostly positive according to the LabMT score.
      Panel F: The function words in the expert-curated LIWC dictionary are not emotionally neutral.
  }
  \label{fig:labMT-tables}
\end{figure*}

In Fig~\ref{fig:main}, we show a direct comparison between word
scores for each pair of the 6 dictionaries tested.
Overall, we find strong agreement between all dictionaries 
with exceptions we note below.
As a guide, we will provide more detail on the individual comparison
between the LabMT dictionary and the other five dictionaries by
examining the words whose scores disagree across dictionaries shown in Fig~\ref{fig:labMT-tables}.
We refer the reader to the S2 Appendix for the remaining individual comparisons.

To start with, consider the comparison the LabMT and ANEW on a word for word basis.
Because these dictionaries share the same range of values, a scatterplot is the
natural way to visualize the comparison.
Across the top row of Fig~\ref{fig:main}, which compares LabMT to the
other 5 dictionaries, we see in Panel B for the LabMT-ANEW comparison 
that the RMA best fit~\cite{rayner1985a} is
\begin{equation*}
  h_{w_{\textnormal{LabMT}}} = 0.92*h_{w_{\textnormal{ANEW}}} + 0.40
\end{equation*}
for words $w_{\textnormal{LabMT}}$ in LabMT and words
$w_{\textnormal{ANEW}}$ in ANEW.
The 10 words with farthest from the line of best fit shown in Panel B of
Fig~\ref{fig:labMT-tables} are, with LabMT and ANEW scores: 
lust (4.64, 7.12), 
bees (5.60, 3.20), 
silly (5.30, 7.41), 
engaged (6.16, 8.00), 
book (7.24, 5.72), 
hospital (3.50, 5.04), 
evil (1.90, 3.23), 
gloom (3.56, 1.88),
anxious (3.42, 4.81), 
and flower (7.88, 6.64).
These are words whose
individual ranges have high standard deviations in LabMT.
While the overall agreement is very good, we should expect some variation
in the emotional associations of words, due to chance, time of survey, and
demographic variability.
Indeed, the Mechanical Turk users who scored the words for the LabMT
set in 2011
are evidently different from the University of Florida
students who took the ANEW survey before 2000 as a class requirement for
Introductory Psychology.

Comparing LabMT with WK in Panel C of Fig~\ref{fig:main}, we again
find a fit with slope near 1, and a smaller positive shift:
$h_{w_{\textnormal{LabMT}}} = 0.96*h_{w_{\textnormal{WK}}}+0.26$.
The 10 words farthest from this line, shown in Panel B of
Fig~\ref{fig:labMT-tables}, are (LabMT, WK): sue (4.30, 2.18), boogie
(5.86, 3.80), exclusive (6.48, 4.50), wake (4.72, 6.57), federal
(4.94, 3.06), stroke (2.58, 4.19), gay (4.44, 6.11), patient (5.04,
6.71), user (5.48, 3.67), and blow (4.48,
6.10).
Like LabMT, the WK dictionary used a Mechanical Turk online survey to
gather word ratings.
We speculate that the minor variation is due in part to the low number
of scores required for each word in the WK survey, with as few as 14
ratings per words and 18 ratings for the majority of the words.
By contrast, LabMT scores represent 50 ratings of each word.
For an in depth comparison, see reference~\cite{dodds2015b}.

Next, in comparing binary dictionaries with $\pm 1$ or $\pm 1,0$ scores to one with a 1--9
range, we can look at the distribution of scores within the continuum score
dictionary for each score in the binary
dictionary.
Looking at the LabMT-MPQA comparison in Panel D of
Fig~\ref{fig:main}, we see that most of the matches are between words
without stems (blue histograms), and that each score in -1, 0, +1 from MPQA corresponds
to a distribution of scores in LabMT.
To examine deviations, we take the words from LabMT sorted by happiest when MPQA is -1, both the happiest and the least happy when MPQA
is 0, and the least happy when MPQA is 1 (Fig~\ref{fig:labMT-tables}
Panels C-E).
The 10 happiest words in LabMT matched by MPQA words with score -1
are: moonlight (7.50), cutest (7.62), finest (7.66), funniest (7.76),
comedy (7.98), laughs (8.18), laughing (8.20), laugh (8.22), laughed
(8.26), laughter (8.50).
This is an immediately troubling list of evidently positive words
somehow rated as -1 in MPQA.
We also see that the top 5 are matched by
the stem ``laugh*'' in MPQA.
The least happy 5 words and happiest 5 words in LabMT matched by words in
MPQA with score 0 are: sorrows (2.69), screaming (2.96), couldn't
(3.32), pressures (3.49), couldnt (3.58), and baby (7.28), precious
(7.34), strength (7.40), surprise (7.42), song (7.58).
Again, we see MPQA word scores are questionable.
The least happy words in LabMT with score +1 in MPQA that are matched by
MPQA are: vulnerable (3.34), court (3.78), sanctions (3.86), defendant
(3.90), conviction (4.10), backwards (4.22), courts (4.24), defendants
(4.26), court's (4.44), and correction (4.44).
Clearly, these words are not positive words in most contexts.

While it would be simple to correct these ratings in the MPQA
dictionary going forward, we have are naturally led to be 
concerned about existing work using MPQA.
We note again that the use of word shifts of some kind would have exposed these problematic
scores immediately.

For the LabMT-LIWC comparison in Panel E of Fig~\ref{fig:main} we
examine the same matched word lists as before.
The 10 happiest words in LabMT matched by words in LIWC with score -1
are: trick (5.22), shakin (5.29), number (5.30), geek (5.34), tricks
(5.38), defence (5.39), dwell (5.47), doubtless (5.92), numbers
(6.04), shakespeare (6.88).
From Panel F of Fig~\ref{fig:labMT-tables}, the least happy 5 neutral
words and happiest 5 neutral words in LIWC, matched with LIWC, are:
negative (2.42), lack (3.16), couldn't (3.32), cannot (3.32), never
(3.34), millions (7.26), couple (7.30), million (7.38), billion
(7.56), millionaire (7.62).
The least happy words in LabMT with score +1 in LIWC that are matched by
LIWC are: merrill (4.90), richardson (5.02), dynamite (5.04), careful
(5.10), richard (5.26), silly (5.30), gloria (5.36), securities
(5.38), boldface (5.40), treasury's (5.42).
The +1 and -1 words in LIWC match some neutral words in LabMT, which
is not alarming.
However, the problems with the ``neutral'' words in the LIWC set are
immediate: these are not emotionally neutral words.
The range of scores in LabMT for these 0-score words in LIWC formed
the basis for Garcia \etal's response to~\cite{dodds2015a}, and we
point out here that the authors must have not looked at the words, and
all-too-common problem in studies using sentiment analysis~\cite{garcia2015a,dodds2015b}.

For the LabMT-OL comparison in Panel E of Fig~\ref{fig:main} we again examine the same matched word lists as before, except the neutral word list because OL has no explicit neutral words.
The 10 happiest words in LabMT matched by OL's negative list are: myth (5.90), puppet (5.90), skinny (5.92), jam (6.02), challenging (6.10), fiction (6.16), lemon (6.16), tenderness (7.06), joke (7.62), funny (7.92).
The least happy words in LabMT with score +1 in OL that are matched by OL are: defeated (2.74), defeat (3.20), envy (3.33), obsession (3.74), tough (3.96), dominated (4.04), unreal (4.57), striking (4.70), sharp (4.84), sensitive (4.86).
Despite nearly twice as many negative words in OL as positive words
(at odds with the frequency-dependent positivity bias of language~\cite{dodds2015a}), these dictionaries generally agree.

\subsection{New {Y}ork {T}imes Word Shift Analysis}
\label{subsec:NYTwordshift}

The New York Times corpus~\cite{nytimescorpus2008a} is split into 
24 sections of the newspaper that are roughly contiguous throughout
the data from 1987--2008.
With each dictionary, we rate each section and then compute word
shifts (described below) against the baseline, and produce a 
happiness ranked list of the
sections.
In the first Figure in S4 Appendix we show scatterplots for each comparison, and compute the Reduced Major Axes (RMA) regression fit~\cite{rayner1985a}.
In the second Figure in S4 Appendix we show the sorted bar chart from each dictionary.

To gain understanding of the sentiment expressed by any given text
relative
to another text, 
it is necessary to inspect the words which contribute most
significantly
by their emotional strength and the change in frequency of usage.
We do this through the use of word shift graphs, which plot the
contribution of each word $w_i$ from the dictionary (denoted $\delta
h_\textnormal{avg} (w_i)$) to the shift in average happiness between
two texts, sorted by the absolute value of the contribution.
We use word shift graphs to both analyze a single text and to
compare two texts, here focusing on comparing text within corpora.
For a derivation of the algorithm used to make word shift graphs while
separating the frequency and sentiment information, we refer the
reader to Equations 2 and 3 in~\cite{dodds2011a}.
We consider both the sentiment difference and frequency difference
parts of $\delta h_\textnormal{avg} (w_i)$ by writing each term of
Eq. \ref{eq:havg} as in~\cite{dodds2011a}:
\begin{multline}
  \delta h_\textnormal{avg} (w_i) 
  =
  \\ 100 \frac{h_\textnormal{avg} (w_i) - h_\textnormal{avg}
    (T_\textnormal{ref})}{h_\textnormal{avg} (T_\textnormal{comp}) -
    h_\textnormal{avg} (T_\textnormal{ref})}  
  \left [ p_i(T_\textnormal{comp}) - p_i (T_\textnormal{ref}) \right ].
\end{multline}
An in-depth explanation of how to interpret the word shift graph can
also be found at
\url{http://hedonometer.org/instructions.html#wordshifts}.

\begin{figure*}[tbp!]
  \includegraphics[width=0.98\textwidth]{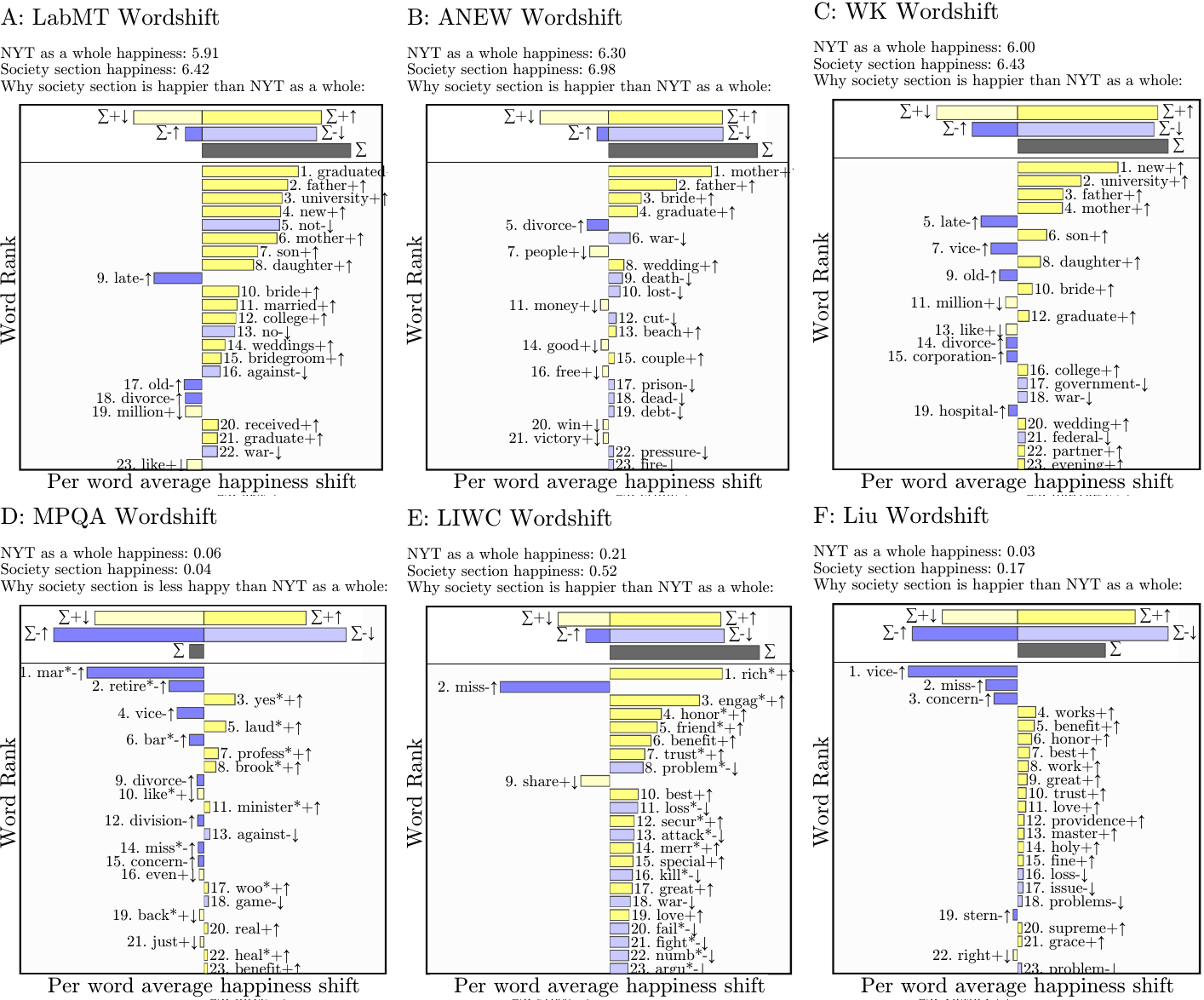}
  \caption{ 
    New York Times (NYT) ``Society'' section shifted against the entire NYT corpus for each of the six dictionaries listed in tiles A--F.
    We provide a detailed analysis in
    Sec. \ref{subsec:NYTwordshift}.
    Generally, we are able to glean the greatest understanding of the
    sentiment texture associated with this NYT section using the LabMT
    dictionary.
    Additionally we note the LabMT dictionary has the most coverage
    quantified by word match count (Figure in S3 Appendix), we
    are able to identify and correct problematic words scores in the
    OL dictionary, and we see that the MPQA dictionary disagrees
    entirely with the others because of an overly broad stem match.
  }
  \label{fig:nyt_wordshifts}
\end{figure*}

To both demonstrate the necessity of using word shift graphs in
carrying out sentiment analysis, and to gain understanding about the
ranking of New York Times sections by each dictionary, we look at word
shifts for the ``Society'' section of the newspaper from each
dictionary in Fig~\ref{fig:nyt_wordshifts}, with the reference text
being the whole of the New York Times.
The ``Society'' section happiness ranks 1, 1, 1, 18, 1, and 11 within the happiness of each of the 24
sections in the dictionaries LabMT, ANEW, WK, MPQA, LIWC, and OL,
respectively.
These shifts show only the very top of the distributions which range
in length from 1030 (ANEW) to 13915 words (WK).

First, using the LabMT dictionary, we see that the words
1. ``graduated'', 2. ``father'', and 3. ``university'' top the list,
which is dominated by positive words that occur more frequently.
These more frequent positive words paint a clear picture of family life
(relationships, weddings, and divorces), as well as university
accomplishment (graduations and college).
In general, we are able to observe with only these words that the
``Society'' section is where we find the details of these positive
events.

From the ANEW dictionary, we see that a few positive words are up,
lead by 1. ``mother'', 2. ``father'', and 3. ``bride''.
Looking at this shift in isolation, we see only these words with three
more (``graduate'', ``wedding'', and ``couple'') that would lead us to
suspect these events are at least common in the ``Society'' section.

The WK dictionary, with the most individual word scores of any
dictionary tested, agrees with LabMT and ANEW that the ``Society''
section is number 1, with somewhat similar set of words at the top:
1. ``new'', 2. ``university'', and 3. ``father''.
Less coverage of the New York Times corpus (see
Fig~\ref{fig:coverage_nyt}) results in the top of the shift showing
less of the character of the ``Society'' section than LabMT, with more
words that go down in frequency in the shift.
With the words ``bride'' and ``wedding'' up, as well as
``university'', ``graduate'', and ``college'', we glean that the
``Society'' section covers both graduations and weddings, as we have
seen so far.

The MPQA dictionary ranks the ``Society'' section 18th of the 24 NYT
sections, a complete departure from the other rankings, with the words
1. ``mar*'', 2. ``retire*'', and 3. ``yes*'' the top three
contributing words.
Negative words increasing in frequency are the most common type near
the top, and of these, the words with the biggest contributions are
being scored incorrectly in this context (specifically words
1. ``mar*'', 2. ``retire*'', 6. ``bar*'', 12. ``division'', and
14. ``miss*'').
Looking more in depth at the problems created by the first of these,
we find 1211 unique words match ``mar*'' with the five most frequent
being married (36750), marriage (5977), marketing (5382), mary (4403),
and mark (2624).
The score for these words in, for example, LabMT are 6.76, 6.7, 5.2,
5.88, and 5.48, confirming our suspicion about these words being
categorized incorrectly with a broad stem match.
These problems plague the MPQA dictionary for scoring the New York
Times corpus, and without using word shifts would have gone completely
unseen.
In an attempt to fix contextual issues by blocking corpus-specific
words, we block ``mar*,retire*,vice,bar*,miss*'' and find that the MPQA dictionary ranks the Society section of the NYT at 15th of the 24 sections

\begin{figure*}[tbp!]
  \includegraphics[width=0.96\textwidth]{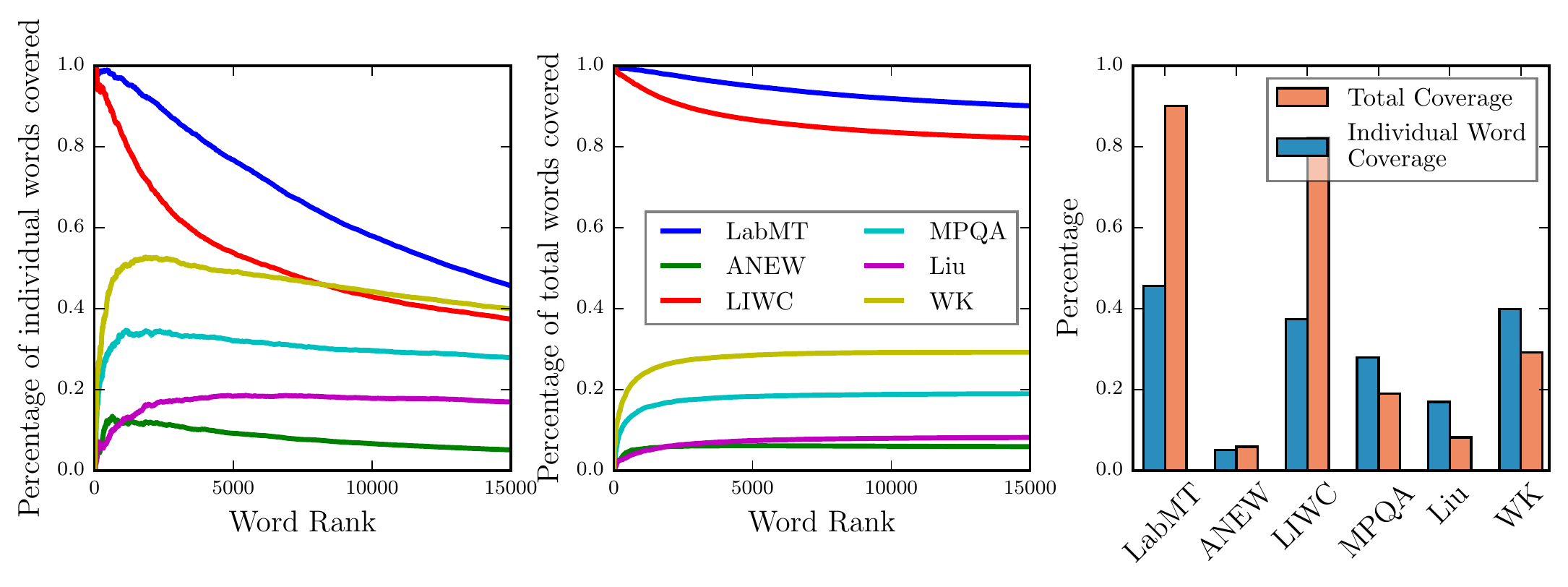}
    \caption{
      Coverage of the words in the movie reviews by each of the dictionaries.
      We observe that the LabMT dictionary has the highest coverage of
      words in the movie reviews corpus both across word rank and
      cumulatively.
      The LIWC dictionary has initially high coverage since it
      contains some high-frequency function words, but quickly drops
      off across rank.
      The WK dictionary coverage increases across word rank and
      cumulatively, indicating that it contains a large number of less
      common words in the movie review corpus.
      The OL, ANEW, and MPQA have a cumulative coverage of less than
      20\% of the lexicon.
    }
  \label{fig:coverage_movies}
\end{figure*}

The second $\pm 1$ dictionary, LIWC, agrees well with the first three
dictionaries and places the ``Society'' section at the top with the
words 1. ``rich*'', 2. ``miss'', and 3. ``engage*'' at the head of the list.
We immediately notice that the word ``miss'' is being used
frequently in the ``Society'' section in a different context than was
rated LIWC: it is used in the corpus to mean the title prefixed to the
name of an unmarried woman, but is scored as negative in LIWC as
meaning to fail to reach an target or to acknowledge loss.
We would remove this word from LIWC for further analysis of this
corpus (we would also remove the word ``trust'' here).
The words matched by ``miss*'' aside, LIWC finds some positive words
going up, with ``engage*'' hinting at weddings.
Otherwise, without words that capture the specific behavior happening
in the ``Society'' section, we are unable to see anything about
college, graduations, or marriages, and there is much less to be
gained about the text from the words in LIWC than some of the other
dictionaries we have seen.
Without these words, it is confirming that LIWC still finds the
``Society'' section to be the top section, due in large part to a lack
of negative words 18. ``war'' and 21. ``fight*''.

The final dictionary from OL disagrees with the others and puts the
``Society'' section at 11th out of the 24 sections.
The top three words, 1. ``vice'', 2. ``miss'', and 3. ``concern'',
contribute largely with respect to the rest of distribution, of which
two are clearly being used in an inappropriate context.
For a more reasonable analysis we would remove both ``vice'' and
``miss'' from the OL dictionary to score this text, making the
``Society'' section the second happiest of the 24 sections.
With this fix, the OL dictionary ranks the Society section of the NYT as the happiest section.
Focusing on the words, we see that the OL dictionary finds many
positive words increasing in frequency that are mostly generic.
In the word shift we do not find the wedding or university events as
in dictionaries with more coverage, but rather a variety of positive
language surrounding these events, for example 4. ``works'',
5. ``benefit'', 6. ``honor'', 7. ``best'', 9. ``great'',
10. ``trust'', 11. ``love'', etc.

In conclusion, we find that 4 of the 6 dictionaries score the
``Society'' section at number 1, and in these cases we use the word
shift to uncover the nuances of the language used.
We find, unsurprisingly, that the most matches are found by the LabMT
dictionary, which is in part built from the NYT corpus
(see S3 Appendix for coverage plots).
Without as much corpus-specific coverage, we note that while the
nuances of the text remain hidden, the LIWC and OL dictionaries still
find the positivity surrounding these unknown events.
Of the two that did not score the ``Society'' section at the top, we repair the MPQA and the OL dictionaries by removing the words ``mar*,retire*,vice*,bar*,miss*'' and ``vice,miss'', respectively.
By identifying words used in the wrong context using the word shift graph, we directly improve the sentiment score for the New York Times corpus from both MPQA and OL dictionaries.
While the OL dictionary, with two corrections, agrees with the other dictionaries, the MPQA dictionary with five corrections still ranks the Society section of the NYT as the 15th happiest of the 24 sections.

\subsection{Movie {R}eviews Classification and Word Shift Analysis}
\label{subsec:moviereviews}

\begin{figure*}[tbp!]
  \centering
  \includegraphics[width=0.98\textwidth]{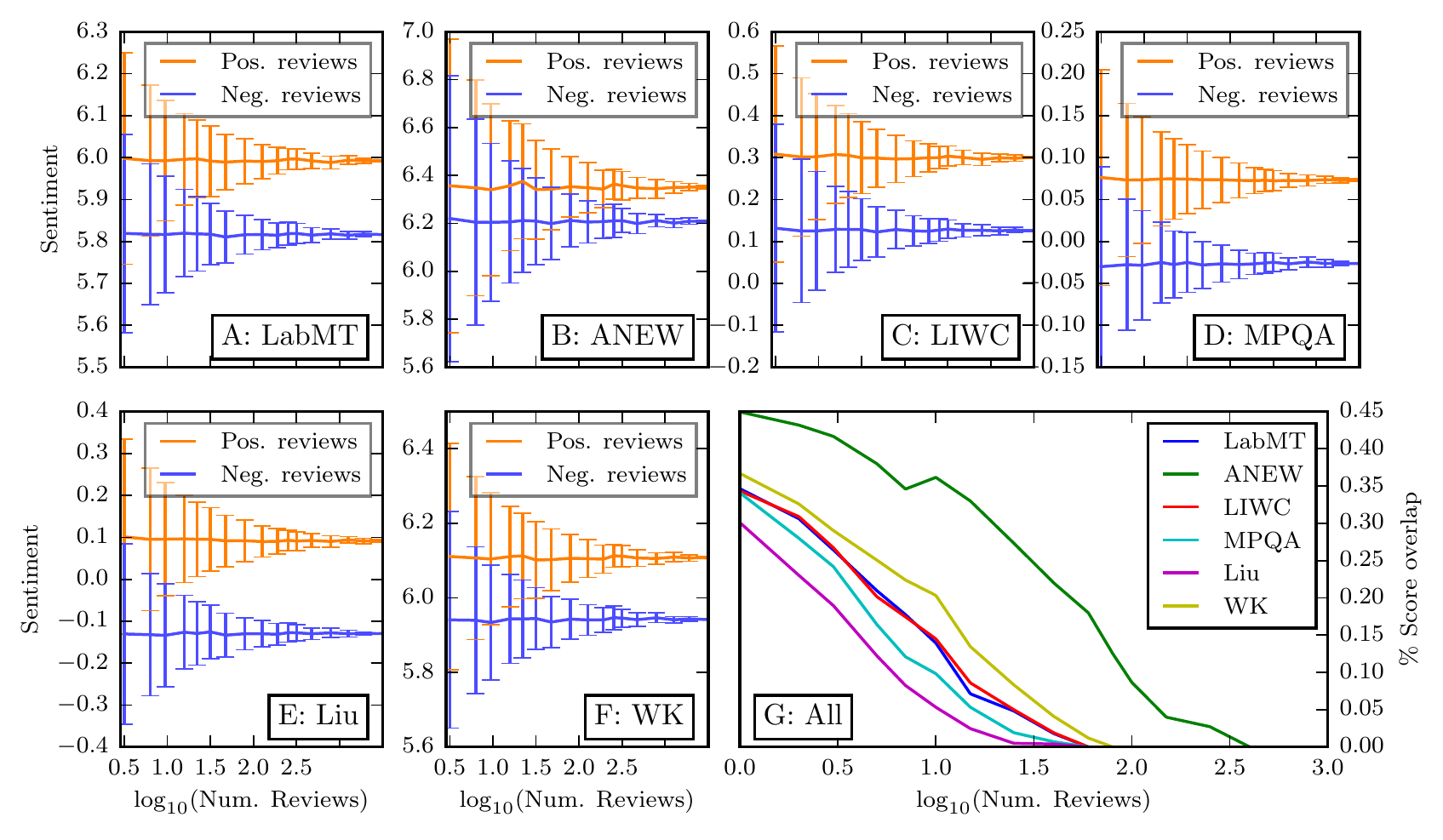}
  \caption{
    The score assigned to increasing numbers of reviews drawn from the tagged positive and negative sets.
    For each dictionary we show mean sentiment and 1 standard
    deviation over 100 samples for each distribution of reviews in
    Panels A--F.
    For comparison we compute the fraction of the distributions that
    overlap in Panel G.
    At the single review level for each dictionary this simple
    performance statistic (fraction of distribution overlap) ranks the
    OL dictionary in first place, the MPQA, LIWC, and LabMT
    dictionaries in a second place tie, WK in fifth, and ANEW far
    behind.
    All dictionaries require on the order of 1000 words to achieve
    95\% classification accuracy.
  }
  \label{fig:moviereviewtest}
\end{figure*}

For the movie reviews, we test the ability to discern positive and
negative reviews.
The entire dataset consists of 1000 positive and 1000 negative
reviews, as rated with 4 or 5 stars and 1 or 2 stars, respectively.
We show how well each dictionary covers the review database
in Fig~\ref{fig:coverage_movies}.
The average review length is 650 words, and we plot the distribution
We average the sentiment of words in each review individually, using
each dictionary.
We also combine random samples of $N$ positive or $N$ negative reviews
for $N$ varying from 2 to 900 on a logarithmic scale,
without replacement, and rate the combined text.
With an increase in the size of the text, we expect that the
dictionaries will be better able to distinguish positive from
negative.
The simple statistic we use to describe this ability is the percentage of
distributions that overlap the average.

In the lower right panel of Fig~\ref{fig:moviereviewtest}, the
percentage overlap of positive and negative review distributions
presents us with a simple summary of dictionary performance on this
tagged corpus.
The ANEW dictionary stands out as being considerably less capable of
distinguishing positive from negative.
In order, we then see WK is slightly better overall, LabMT and LIWC
perform similarly better than WK overall, and then MPQA and OL are
each a degree better again, across the review lengths (see below for
hard numbers at 1 review length).
Two Figures in S5 Appendix
show the distributions for 1 review and for 15 combined reviews.

\begin{figure*}[tbp!]
  \includegraphics[width=0.98\textwidth]{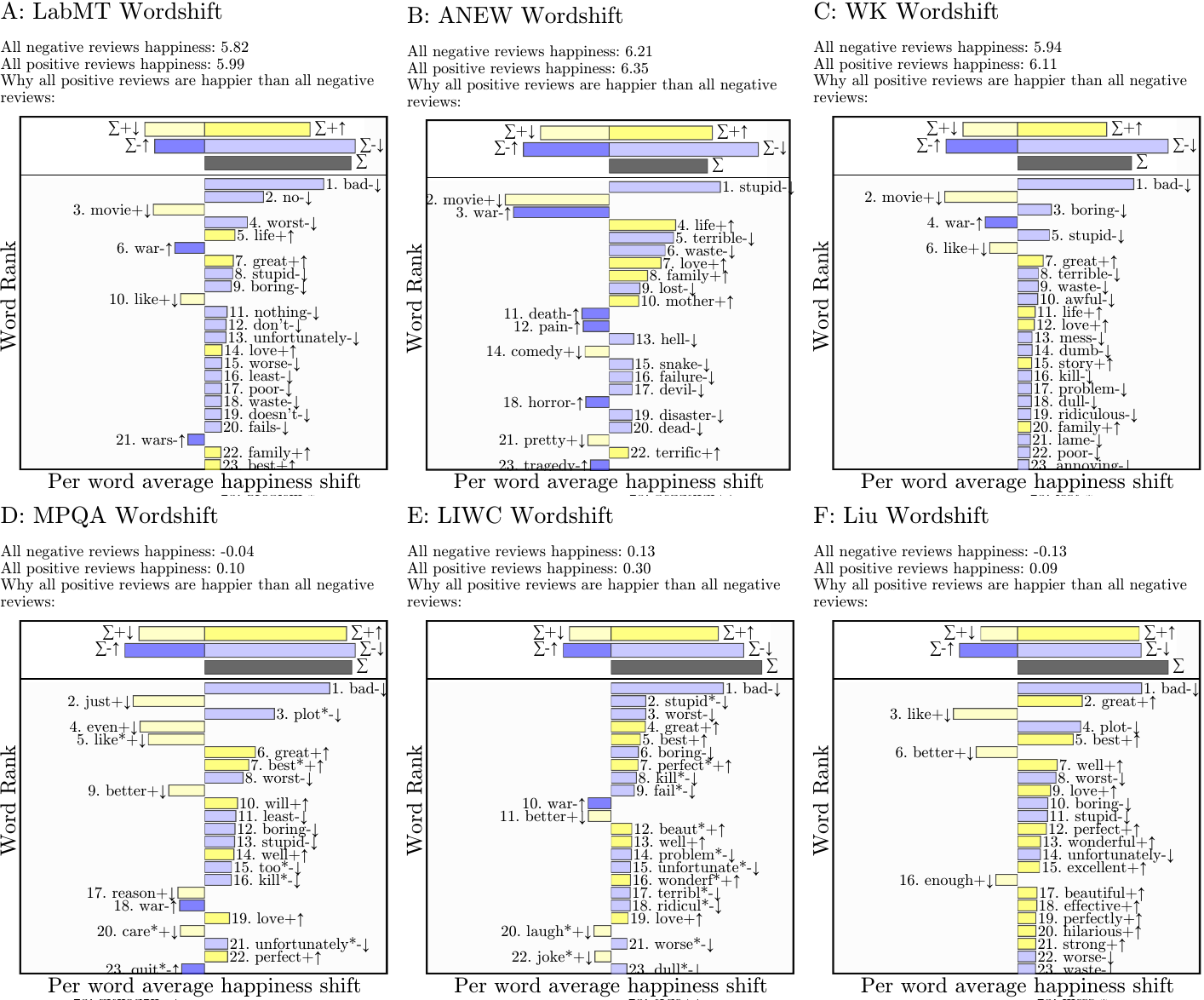}
  \caption{
    Word shifts for the movie review corpus.
    By analyzing the words that contribute most significantly to the
    sentiment score produced by each dictionary we are able to
    identify words that are problematic for each dictionary at the
    word-level, and generate an understanding of the emotional texture
    of the movie review corpus.
    Again we find that coverage of the lexicon is essential to produce
    meaningful word shifts, with the LabMT dictionary providing the
    most coverage of this corpus and producing the most detailed word
    shifts.
    All words on the left hand side of these word shifts are words
    that individually made the positive reviews score more negatively than the
    negative reviews, and the removal of these words would improve the
    accuracy of the ratings given by each dictionary.
    In particular, across each dictionary the word shifts show that
    domain-specific words such as ``war'' and ``movie'' are used more
    frequently in the positive reviews and are not useful in
    determining the polarity of a single review.
  }
  \label{fig:moviereviews-shifts}
\end{figure*}

To analyze which words are being used by each dictionary, we compute
word shift graphs of the entire positive corpus versus the entire
negative corpus in Fig~\ref{fig:moviereviews-shifts}.
Across the board, we see that a decrease in negative words is the most
important word type for each dictionary, with the word ``bad'' being
the top word for every dictionary in which it is scored (ANEW does not
have it).
Other observations that we can make from the word shifts include a few
words that are potentially being used out of context: ``movie'',
``comedy'', ``plot'', ``horror'', ``war'', ``just''.

Classifying single reviews as positive or negative, the F1-scores
are: LabMT .63, ANEW .36, LIWC .53, MPQA .66, OL .71, and WK
.34 (see Table \ref{tbl:MR-1}).
We roughly confirm the rule-of-thumb that 10,000 words are enough to
score with a dictionary confidently, with all dictionaries except MPQA
and ANEW achieving 90\% accuracy with this many words.
We sample the number of reviews evenly in log space, generating sets
of reviews with average word counts of 4550, 6500, 9750, 16250, and
26000 words.
Specifically, the number of reviews necessary to achieve 90\% accuracy
is 15 reviews (9750 words) for LabMT, 100 reviews (65000 words) for
ANEW, 10 reviews (6500 words) for LIWC, 10 reviews (6500 words) for
MPQA, 7 reviews (4550 words) for OL, and 25 reviews (16250 words) for
WK.

The OL dictionary, with the highest performance classifying individual movie reviews of the 6 dictionaries tested in detail, performs worse than guessing at classifying individual sentences in movie reviews.
Specifically, 76.9/74.2\% of sentences in the positive/negative reviews sets have words in the OL dictionary, and of these OL achieves an F1-score of 0.44.
The results for each dictionary are included in Table \ref{tbl:MR-2}, with an average (median) F1 score of 0.42 (0.45) across all dictionaries.

\subsection{Google {B}ooks Time Series and Word Shift Analysis}
\label{subsec:googlebooks}

We use the Google books 2012 dataset with all English
books~\cite{lin2012syntactic}, from which we remove part of speech
tagging and split into years.
From this, we make time series by year, and word shifts of decades
versus the baseline.
In addition, to assess the similarity of each time series, we produce
correlations between each of the time series.

Despite grand claims from research based on the Google Books corpus
\cite{michel2011quantitative}, we keep in mind that there are several
deep problems with this beguiling data set~\cite{pechenick2015a}.
Leaving aside these issues, the Google Books corpus nevertheless 
provides a substantive test of our six dictionaries.

\begin{figure*}[tbp!]
  \includegraphics[width=0.98\textwidth]{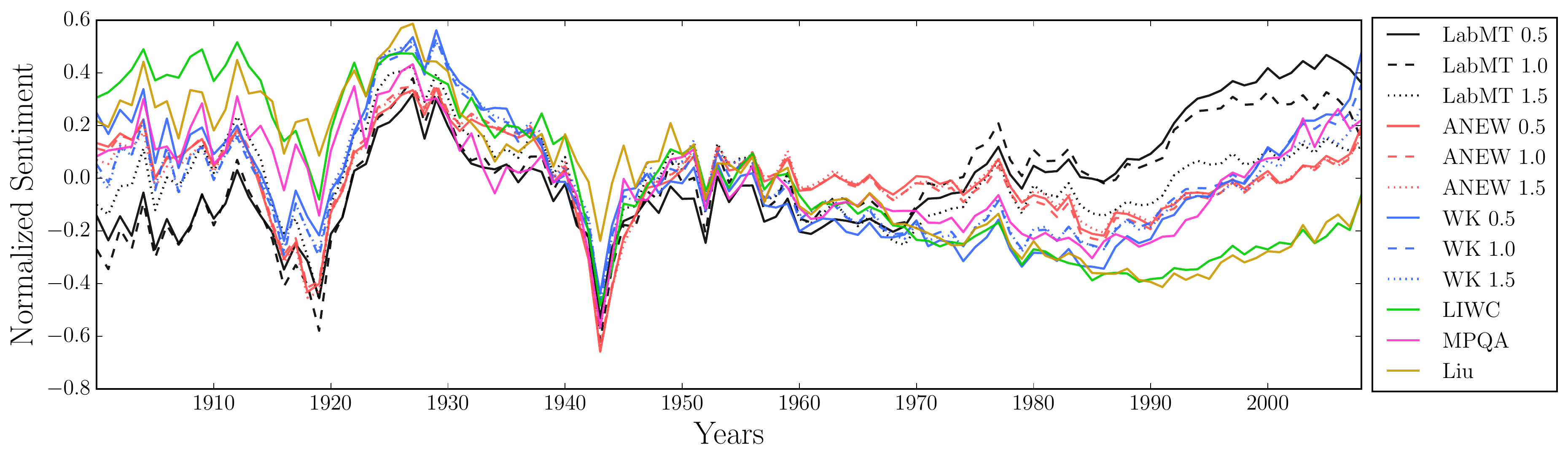}
  \caption{ 
    Google Books sentiment time series from each dictionary, with stop values of 0.5, 1.0, and 1.5 from the dictionaries with word scores in the 1--9 range.
    To normalize the sentiment score, we subtract the mean and divide
    by the absolute range.
    We observe that each time series is choppy, with a few pronounced
    negative time periods, and trending positive towards the end of
    the corpus.
    The score of LabMT varies substantially with different stop
    values, although remaining highly correlated, and finds absolute
    lows near the World Wars.
    The LIWC and OL dictionaries trend down towards 1990, dipping as
    low as the war periods.
  }
  \label{fig:gbooks_timeseries}
\end{figure*}

In Fig~\ref{fig:gbooks_timeseries}, we plot the sentiment time series
for Google Books.  Three immediate trends stand out: a dip near the
Great Depression, a dip near World War II, and a general upswing in
the 1990's and 2000's.  From these general trends, a few dictionaries
waver: OL does not dip very much for WW2, OL and LIWC stay lower in
the 90's and 2000's, and LabMT with $\Delta_h = 0.5,1.0$ go downward
near the end of the 2000's.  We take a closer look into the 1940's to
see what each dictionary is picking up in Google Books around World
War 2 in Figure in S6 Appendix.

In each panel of the word shift Figure in S6 Appendix, we see that the top word making the 1940's less positive than the the rest of Google Books is ``war'', which is the top contributor for every dictionary except OL.
Rounding out the top three contributing words are ``no'' and ``great'', and we infer that the word ``great'' is being seen from mention of ``The Great Depression'' or ``The Great War'', and is possibly being used out of context.
All dictionaries but ANEW have ``great'' in the top 3 words, and each dictionary could be made more accurate if we remove this word.

In Panel A of the 1940's word shift Figure in S6 Appendix, beyond the top words, increasing words are mostly negative and war-related: ``against'', ``enemy'', ``operation'', which we could expect from this time period.

In Panel B, the ANEW dictionary scores the 1940's of Google Books
lower than the baseline as well, finding ``war'', ``cancer'', and
``cell'' to be the most important three words.
With only 1030 words, there is not enough coverage to see anything
beyond the top word ``war,'' and the shift is dominated by words that
go down in frequency.

In Panel C, the WK dictionary finds the the 1940's with slightly less happy than the baseline, with the top three words being ``war'', ``great'', and ``old''.
We see many of the same war-related words as in LabMT, and in addition some positive words like ``good'' and ``be'' are up in frequency.
The word ``first'' could be an artifact of first aid.

In Panel D, the MPQA dictionary rates the 1940's with slightly less happy than the baseline, with the top three words being ``war'', ``great'', and ``differ*''.
Beyond the top word ``war'', the score is dominated by words decreasing in frequency, with only a few words up in frequency.
Without specific words being up in frequency, it is difficult to obtain a good glance at the nature of the text here.

In Panel E, the LIWC dictionary rates the 1940's nearly the same as the baseline, with the top three words being ``war'', ``great'', and ``argu*''.
When the scores are nearly the same, although the 1940's are slightly higher happiness here, the word shift is a view into how the words of the reference and comparison text vary.
In addition to a few war related words being up and bringing the score down (``fight'', ``enemy'', ``attack''), we see some positive words up that could also be war related: ``certain'', ``interest'', and ``definite''.
Although LIWC does not manage to find World War II as a low point of the 20th century, the words that it generates are useful in understanding the corpus.

In Panel F, the OL dictionary rates the 1940's as happier than the baseline, with the top three words being ``great'', ``support'', and ``like''.
With 7 positive words up, and 1 negative word up, we see how the OL dictionary misses the war without the word ``war'' itself and with only ``enemy'' contributing from the words surrounding the conflict.
The nature of the positive words that are up is unclear, and could justify a more detailed analysis of why the OL dictionary fails here.

\subsection{Twitter Time Series Analysis}
\label{subsec:twittertimeseries}

We store data on the Vermont Advanced Computing Core (VACC), and process the text first into hash tables (with approximately 8 million unique English words each day) and then into word vectors for each 15 minutes, for each dictionary tested.
From this, we build sentiment time series for time resolutions of 15 minutes, 1 hour, 3 hours, 12 hours, and 1 day.
In addition to the raw time series, we compute correlations between each time series to assess the similarity of the ratings between dictionaries.

\begin{figure*}[tbp!]
  \centering
  \includegraphics[width=0.98\textwidth]{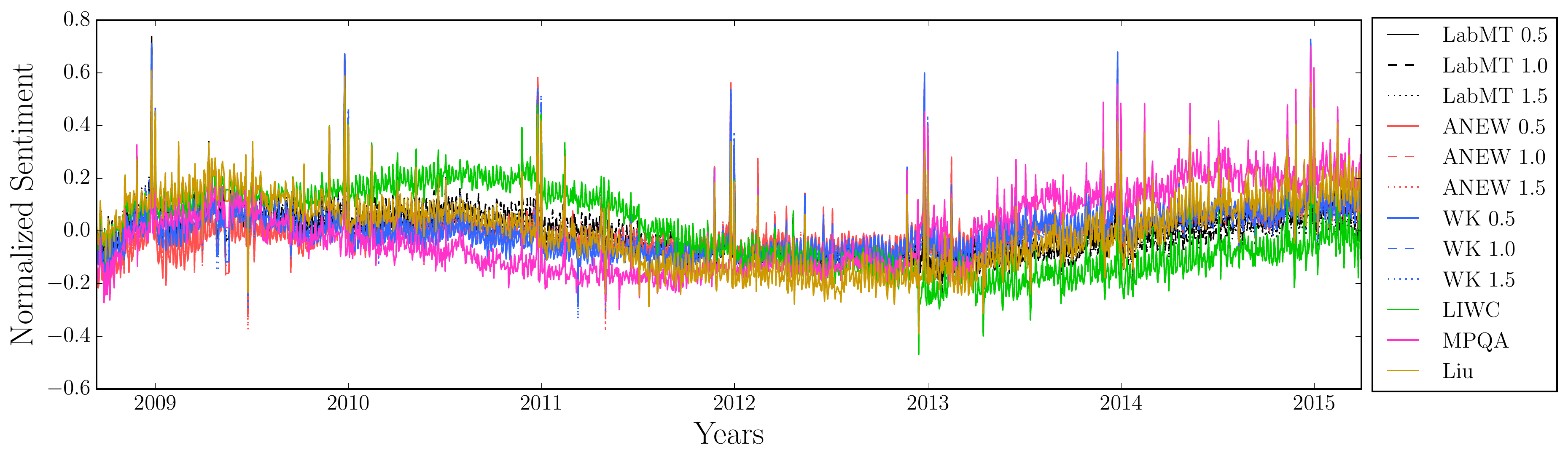}
  \caption{
    Normalized sentiment time series on Twitter using $\Delta _h$ of 1.0 for all dictionaries.
    To normalize the sentiment score, we subtract the mean and divide by the absolute range.
    The resolution is 1 day, and draws on the 10\% gardenhose sample of public tweets provided by Twitter.
    All of the dictionaries exhibit wide variation for very early Tweets, and from 2012 onward generally track together strongly with the exception of MPQA and LIWC.
    The LIWC and MPQA dictionaries show opposite trends: a rise until 2012 with a decline after 2012 from LIWC, and a decline before 2012 with a rise afterwards from MPQA.
        To analyze the trends we look at the words driving the movement across years using word shift Figures in S7 Appendix.
    }
  \label{fig:twitter_timeseries_4}
\end{figure*}

In Fig~\ref{fig:twitter_timeseries_4}, we present a daily sentiment time series of Twitter processed using each of the dictionaries being tested.
With the exception of LIWC and MPQA we observe that the dictionaries generally track well together across the entire range.
A strong weekly cycle is present in all, although muted for ANEW.

\begin{figure}[tbp!]
  \centering
  \includegraphics[width=0.48\textwidth]{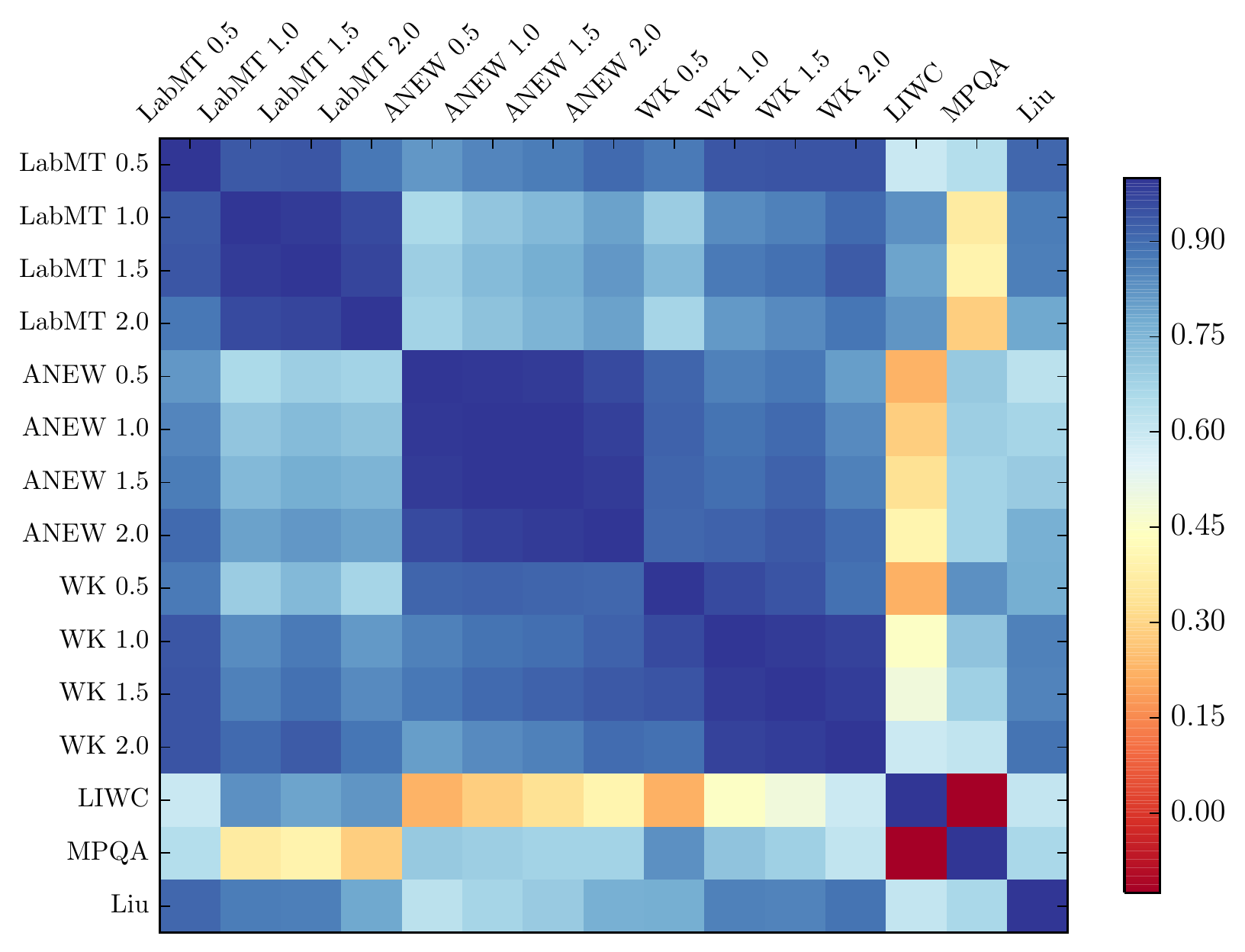}
  \caption{
    Pearson's $r$ correlation between daily resolution Twitter sentiment time series for each dictionary.
    We see that there is strong agreement within dictionaries, with
    the biggest differences coming from the stop value of $\Delta h =
    0.5$.
    The LabMT and OL dictionaries do not strongly disagree with any
    of the others, while LIWC is the least correlated overall with
    other dictionaries.
    LabMT and OL correlate strongly with each other, and disagree
    most with the ANEW, LIWC, and MPQA dictionaries.
    The two least correlated dictionaries are the LIWC and MPQA
    dictionaries.
    Again, since there is no publicly accessible ground truth for Twitter sentiment, we
    compare dictionaries against the others, and look at the words.
    With these criteria, we find the LabMT dictionary to be the most
    useful.
  }
  \label{fig:twitter_correlation_4}
\end{figure}

We plot the Pearson's correlation between all time series in Fig~\ref{fig:twitter_correlation_4}, and confirm some of the general observations that we can make from the time series.
Namely, the LIWC and MPQA time series disagree the most from the others, and even more so with each other.
Generally, we see strong agreement within dictionaries with varying stop values $\Delta h$.

All of the dictionaries are choppy at the start of the time frame,
when Twitter volume is low in 2008 and into 2009.
As more people join Twitter and the tweet volume increases through
2010, we see that LIWC rates the text as happier, while the rest start a slow
decline in rating that is led by MPQA in the negative direction.
In 2010, the LIWC dictionary is more positive than the rest with words
like ``haha'', ``lol'' and ``hey'' being used more frequently and
swearing being less frequent than the all years of Twitter put
together.
The other dictionaries with more coverage find a decrease in positive
words to balance this increase, with the exception of MPQA which finds
many negative words going up in frequency (see 2010 word shift Figure in Appendix S7).
All of the dictionaries agree most strongly in 2012, all finding a lot
of negative language and swearing that brings scores down 
(see 2012 word shift Figure in Appendix S7).
From the bottom at 2012, LIWC continues to go downward while the
others trend back up.
The signal from MPQA jumps to the most positive, and LIWC does start
trending back up eventually.
We analyze the words in 2014 with a word shift against all 7 years of
tweets for each dictionary in each panel in the
2014 word shift Figure in Appendix S7:
A. LabMT finds 2014 with less happy with
more negative language. B. ANEW finds it happier with a few
positive words up. C. WK finds it happier with more negative words
(like LabMT). D. MPQA finds it more positive with less negative
words. E. LIWC finds it less positive with more negative and less
positive words. F. OL finds it to be of the same sentiment as the
background with a balance in positive and negative word usage.
From these word shifts, we can analyze which words cause MPQA and LIWC
to disagree with the other dictionaries: the disagreement of MPQA is
again marred by broad stem matches, and the disagreement of LIWC is
due to a lack of coverage.

\subsection{Brief Comparison to Machine Learning Methods}
\label{subsec:NB-section}

We implement a Naive Bayes (NB) classifier (sometimes harshly called idiot Bayes~\cite{hand2001idiot}) on the tagged movie review dataset, using 10\% of the reviews for training and then testing performance on the rest.
Following standard practice, we remove the top 30 ranked words (``stop words'') from the 5000 most frequent words, and use the remaining 4970 words in our classifier for maximum performance (we observe a 0.5\% improvement).
Our implementation is analogous to those found in common Python natural language processing packages (see ``NLTK'' or ``TextBlob'' \cite{bird2006nltk}).

As we should expect, at the level of single review, NB outperforms the dictionary-based methods with a classification accuracy of 75.7\% averaged over 100 trials.
As the number of reviews is increased, the overlap from NB diminishes, and using our simple ``fraction overlapping'' metric, the error drops to 0 with more than 200 reviews.
Interestingly, NB starts to do worse with more reviews put together, and with more than 500 of the 1000 reviews put together, it rates both the positive and negative reviews as positive (Figure in S8 Appendix).

The rating curves do not touch, and neither do the error bars, but they both go very slightly above 0.
Overall, with Naive Bayes we are able to classify a higher percentage of individual reviews correctly, but with more variance.

In the two Tables in S8 Appendix we compute the words which the NB classifier uses to classify all of the positive reviews as positive, and all of the negative reviews as positive.
The Natural Language Toolkit~\cite{bird2006nltk} implements a method to obtain the ``most informative'' words, by taking the ratio of the likelihood of words between all available classes, and looking for the largest ratio:
\begin{equation}
  \max _{\textnormal{all words }w} \frac{P ( w | c_i )}{P ( w | c_j )}
\end{equation}
for all combinations of classes $c_i,c_j$.
This is possible because of the ``naive'' assumption that feature (word) likelihoods are independent, resulting in a classification metric that is linear for each feature.
In S8 Appendix, we provide the derivation of this linearity structure.

We find that the trained NB classifier relies heavily on words that
are very specific to the training set including the names of actors of
the movies themselves, making them useful as classifiers but not in
understanding the nature of the text.
We report the top 10 words for both positive and negative classes
using both the ratio and difference methods in Table in S8 Appendix.
To classify a document using NB, we use the frequency of each word in the
document in conjunction with the probability that that word
occurred in each labeled class $c_i$.

We next take the movie-review-trained NB classifier and use it to
classify the New York Times sections, both by ranking them and by looking at
the words (the above ratio and difference weighted by the occurrence of
the words).
We ranked the sections 5 different times, and among those 
find the ``Television'' section both by far the happiest, 
and by far the least happy in independent tests.
We show these rankings and report the top 10 words used to score the ``Society'' section in Table~\ref{tbl:NB-2}.

We thus see that the NB classifier, a linear learning method, may perform poorly when assessing
sentiment outside of the corpus on which it is trained.
In general, performance will vary depending on
the statistical dissimilarity of the training and novel corpora.
Added to this is the inscrutability of black box methods:
while susceptible to the aforementioned difficulty, nonlinear learning methods (unlike NB) also render detailed examination of how
individual words contribute to a text's score more difficult.

\section{Conclusion}
\label{sec:conclusion}

We have shown that measuring sentiment in various corpora presents
unique challenges, and that dictionary performance is situation
dependent.
Across the board, the ANEW dictionary performs poorly, and the
continued use of this dictionary with clearly better alternatives is a
questionable choice.
We have seen that the MPQA dictionary does not agree with the other
five dictionaries on the NYT corpus and Twitter corpus due to a
variety of context and stem matching issues, and we would not
recommend using this dictionary.
And in comparison to LabMT, the WK, LIWC, and OL dictionaries fail to
provide much detail in corpora where their coverage is lower,
including all four corpora tested.
Sufficient coverage is essential or producing meaningful word shifts and
thereby enabling deeper understanding.

In each case, to analyze the output of the dictionary method, 
we rely on the use of word shift graphs.
With this tool, we can produce a finer grained analysis of the lexical
content, and we can also detect words that are used out of context and can mask them
directly.
It should be clear that using any of the dictionary-based sentiment detecting method 
without looking at how individual words contribute is indefensible,
and analyses that do not use word shifts or similar tools cannot be trusted.
The poor word shift performance of binary dictionaries in particular
gravely limits their ability to reveal underlying stories.

In sum, we believe that dictionary-based methods will continue
to play a powerful role---they are fast and well suited for web-scale
data sets---and that the best instruments will be based on dictionaries
with excellent coverage and continuum scores.
To this end, we urge that all dictionaries should be regularly updated
to capture changing lexicons, word usage, and demographics.
Looking further ahead, a move from scoring words to scoring 
both phrases and words should realize
considerable improvement for many languages of interest.
With phrase dictionaries, the resulting phrase shift graphs
will allow for a  more nuanced and detailed analysis of a corpus's sentiment
score~\cite{alajajian2015a},
ultimately affording clearer stories 
for sentiment dynamics.

\bibliographystyle{unsrtabbrv}

\clearpage
\pagebreak

\newwrite\tempfile
\immediate\openout\tempfile=startsupp.txt
\immediate\write\tempfile{\thepage}
\immediate\closeout\tempfile

\onecolumngrid

\setcounter{page}{1}
\renewcommand{\thepage}{S\arabic{page}}
\renewcommand{\thefigure}{S\arabic{figure}}
\renewcommand{\thetable}{S\arabic{table}}
\setcounter{figure}{0}
\setcounter{table}{0}

\setcounter{section}{0}

\section*{S1 Appendix: Computational methods} \label{supp:methods}

All of the code to perform these tests is available and document on GitHub.
The repository can be found here: \url{https://github.com/andyreagan/sentiment-analysis-comparison}.

\subsection*{Stem matching} \label{supp:stems}

Of the dictionaries tested, both LIWC and MPQA use ``word stems''.
Here we quickly note some of the technical difficulties with using word stems, and how we processed them, for future research to build upon and improve.

An example is \verb|abandon*|, which is intended to the match words of the standard RE form \verb|abandon[a-z]*|.
A naive approach is to check each word against the regular expression, but this is prohibitively slow.
We store each of the dictionaries in a ``trie'' data structure with a record.
We use the easily available ``marisa-trie'' Python library, which wraps the C++ counterpart.
The speed of these libraries made the comparison possible over large corpora, in particular for the dictionaries with stemmed words, where the \verb|prefix| search is necessary.
Specifically, the ``trie'' structure is 70 times faster than a regular expression based search for stem words.
In particular, we construct two tries for each dictionary: a fixed and stemmed trie.
We first attempt to match words against the fixed list, and then turn to the prefix match on the stemmed list.

\subsection*{Regular expression parsing}

The first step in processing the text of each corpora is extracting the words from the raw text.
Here we rely on a regular expression search, after first removing some punctuation.
We choose to include a set of all characters that are found within the words in each of the six dictionaries tested in detail, such that it respects the parse used to create these dictionaries by retaining such characters.
This takes the following form in Python, for \verb|raw_text| as a string (note, \verb|pdflatex| renders correctly locally, but arXiv seems to explode the link match group):
\begin{verbatim}
punctuation_to_replace = ["---","--","''"]
for punctuation in punctuation_to_replace:
    raw_text = raw_text.replace(punctuation," ")
words = [x.lower() for x in re.findall(r"(?:[0-9][0-9,\.]*[0-9])|
                                         (?:http://[\w\./\-\?\&\#]+)|
                                         (?:[\w\@\#\'\&\]\[]+)|
                                         (?:[b}/3D;p)|’\-@x#^_0\\P(o:O{X$[=<>\]*B]+)",
                                       raw_text,flags=re.UNICODE)]
\end{verbatim}

\clearpage
\pagebreak

\section*{S2 Appendix: Continued individual comparisons} \label{supp:comparisons}

Picking up right where we left off in Section \ref{sec:results}, we next compare ANEW with the other dictionaries.
The ANEW-WK comparison in Panel I of Fig. \ref{fig:main} contains all 1030 words of ANEW, with a fit of $h_{w_{\text{ANEW}}} = 1.07*h_{w_{\text{WK}}}-0.30$, making ANEW more positive and with increasing positivity for more positive words.
The 20 most different scores are (ANEW,WK): fame (7.93,5.45), god (8.15,5.90), aggressive (5.10,3.08), casino (6.81,4.68), rancid (4.34,2.38), bees (3.20,5.14), teacher (5.68,7.37), priest (6.42,4.50), aroused (7.97,5.95), skijump (7.06,5.11), noisy (5.02,3.21), heroin (4.36,2.74), insolent (4.35,2.74), rain (5.08,6.58), patient (5.29,6.71), pancakes (6.08,7.43), hospital (5.04,3.52), valentine (8.11,6.40), and book (5.72,7.05).
We again see some of the same words from the LabMT comparisons with these dictionaries, and again can attribute some differences to small sample sizes and differing demographics.

For the ANEW-MPQA comparison in Panel J of Fig. \ref{fig:main} we show the same matched word lists as before.
The happiest 10 words in ANEW matched by MPQA are: clouds (6.18), bar (6.42), mind (6.68), game (6.98), sapphire (7.00), silly (7.41), flirt (7.52), rollercoaster (8.02), comedy (8.37), laughter (8.45).
The least happy 5 neutral words and happiest 5 neutral words in MPQA, matched with MPQA, are: pressure (3.38), needle (3.82), quiet (5.58), key (5.68), alert (6.20), surprised (7.47), memories (7.48), knowledge (7.58), nature (7.65), engaged (8.00), baby (8.22).
The least happy words in ANEW with score +1 in MPQA that are matched by MPQA are: terrified (1.72), meek (3.87), plain (4.39), obey (4.52), contents (4.89), patient (5.29), reverent (5.35), basket (5.45), repentant (5.53), trumpet (5.75).
Again we see some very questionable matches by the MPQA dictionary, with broad stems capturing words with both positive and negative scores.

For the ANEW-LIWC comparison in Panel K of Fig. \ref{fig:main} we show the same matched word lists as before.
The happiest 10 words in ANEW matched by LIWC are: lazy (4.38), neurotic (4.45), startled (4.50), obsession (4.52), skeptical (4.52), shy (4.64), anxious (4.81), tease (4.84), serious (5.08), aggressive (5.10).
There are only 5 words in ANEW that are matched by LIWC with LIWC score of 0: part (5.11), item (5.26), quick (6.64), couple (7.41), millionaire (8.03).
The least happy words in ANEW with score +1 in LIWC that are matched by LIWC are: heroin (4.36), virtue (6.22), save (6.45), favor (6.46), innocent (6.51), nice (6.55), trust (6.68), radiant (6.73), glamour (6.76), charm (6.77).

For the ANEW-Liu comparison in Panel L of Fig. \ref{fig:main} we show the same matched word lists as before, except the neutral word list because Liu has no explicit neutral words.
The happiest 10 words in ANEW matched by Liu are: pig (5.07), aggressive (5.10), tank (5.16), busybody (5.17), hard (5.22), mischief (5.57), silly (7.41), flirt (7.52), rollercoaster (8.02), joke (8.10).
The least happy words in ANEW with score +1 in Liu that are matched by Liu are: defeated (2.34), obsession (4.52), patient (5.29), reverent (5.35), quiet (5.58), trumpet (5.75), modest (5.76), humble (5.86), salute (5.92), idol (6.12).

For the WK-MPQA comparison in Panel P of Fig. \ref{fig:main} we show the same matched word lists as before.
The happiest 10 words in WK matched by MPQA are: cutie (7.43), pancakes (7.43), panda (7.55), laugh (7.56), marriage (7.56), lullaby (7.57), fudge (7.62), pancake (7.71), comedy (8.05), laughter (8.05).
The least happy 5 neutral words and happiest 5 neutral words in MPQA, matched with MPQA, are: sociopath (2.44), infectious (2.63), sob (2.65), soulless (2.71), infertility (3.00), thinker (7.26), knowledge (7.28), legacy (7.38), surprise (7.44), song (7.59).
The least happy words in WK with score +1 in MPQA that are matched by MPQA are: kidnapper (1.77), kidnapping (2.05), kidnap (2.19), discriminating (2.33), terrified (2.51), terrifying (2.63), terrify (2.84), courtroom (2.84), backfire (3.00), indebted (3.21).

For the WK-LIWC comparison in Panel Q of Fig. \ref{fig:main} we show the same matched word lists as before.
The happiest 10 words in WK matched by LIWC are: geek (5.56), number (5.59), fiery (5.70), trivia (5.70), screwdriver (5.76), foolproof (5.82), serious (5.88), yearn (5.95), dumpling (6.48), weeping willow (6.53).
The least happy 5 neutral words and happiest 5 neutral words in LIWC, matched with LIWC, are: negative (2.52), negativity (2.74), quicksand (3.62), lack (3.68), wont (4.09), unique (7.32), millionaire (7.32), first (7.33), million (7.55), rest (7.86).
The least happy words in WK with score +1 in LIWC that are matched by LIWC are: heroin (2.74), friendless (3.15), promiscuous (3.32), supremacy (3.48), faithless (3.57), laughingstock (3.77), promiscuity (3.95), tenderfoot (4.26), succession (4.52), dynamite (4.79).

For the WK-Liu comparison in Panel R of Fig. \ref{fig:main} we show the same matched word lists as before, except the neutral word list because Liu has no explicit neutral words.
The happiest 10 words in WK matched by Liu are: goofy (6.71), silly (6.72), flirt (6.73), rollercoaster (6.75), tenderness (6.89), shimmer (6.95), comical (6.95), fanciful (7.05), funny (7.59), fudge (7.62), joke (7.88).
The least happy words in WK with score +1 in Liu that are matched by Liu are: defeated (2.59), envy (3.05), indebted (3.21), supremacy (3.48), defeat (3.74), overtake (3.95), trump (4.18), obsession (4.38), dominate (4.40), tough (4.45).

Now we'll focus our attention on the MPQA row, and first we see comparisons against the three full range dictionaries.
For the first match against LabMT in Panel D of Fig. \ref{fig:main}, the MPQA match catches 431 words with MPQA score 0, while LabMT (without stems) matches 268 words in MPQA  in Panel S (1039/809 and 886/766 for the positive and negative words of MPQA).
Since we've already highlighted most of these words, we move on and focus our attention on comparing the $\pm 1$ dictionaries.

In Panels V--X, BB--DD, and HH--JJ of Fig. \ref{fig:main} there are a total of 6 bins off of the diagonal, and we focus out attention on the bins that represent words that have opposite scores in each of the dictionaries.
For example, consider the matches made my MPQA in Panel BB: the words in the top left corner and bottom right corner with are scored in a opposite manner in LIWC, and are of particular concern.
Looking at the words from Panel W with a +1 in MPQA and a -1 in LIWC (matched by LIWC) we see: stunned, fiery, terrified, terrifying, yearn, defense, doubtless, foolproof, risk-free, exhaustively, exhaustive, blameless, low-risk, low-cost, lower-priced, guiltless, vulnerable, yearningly, and yearning.
The words with a -1 in MPQA that are +1 in LIWC (matched by LIWC) are: silly, madly, flirt, laugh, keen, superiority, supremacy, sillily, dearth, comedy, challenge, challenging, cheerless, faithless, laughable, laughably, laughingstock, laughter, laugh, grating, opportunistic, joker, challenge, flirty.

In Panel W of \ref{fig:main}, the words with a +1 in MPQA and a -1 in Liu (matched by Liu) are: solicitude, flair, funny, resurgent, untouched, tenderness, giddy, vulnerable, and joke.
The words with a -1 in MPQA that are +1 in Liu, matched by Liu, are: superiority, supremacy, sharp, defeat, dumbfounded, affectation, charisma, formidable, envy, empathy, trivially, obsessions, and obsession.

In Panel BB of \ref{fig:main}, the words with a +1 in LIWC and a -1 in MQPA (matched by MPQA) are: silly, madly, flirt, laugh, keen, determined, determina, funn, fearless, painl, cute, cutie, and gratef.
The words with a -1 in LIWC and a +1 in MQPA, that are matched by MPQA, are: stunned, terrified, terrifying, fiery, yearn, terrify, aversi, pressur, careless, helpless, and hopeless.

In Panel DD of \ref{fig:main}, the words with a -1 in LIWC and a +1 in Liu, that are matched by Liu, are: silly, and madly.
The words with a +1 in LIWC and a -1 in Liu, that are matched by Liu, are: stunned, and fiery.

In Panel HH of \ref{fig:main}, the words with a -1 in Liu and a +1 in MPQA, that are matched by MPQA, are: superiority, supremacy, sharp, defeat, dumbfounded, charisma, affectation, formidable, envy, empathy, trivially, obsessions, obsession, stabilize, defeated, defeating, defeats, dominated, dominates, dominate, dumbfounding, cajole, cuteness, faultless, flashy, fine-looking, finer, finest, panoramic, pain-free, retractable, believeable, blockbuster, empathize, err-free, mind-blowing, marvelled, marveled, trouble-free, thumb-up, thumbs-up, long-lasting, and viewable.
The words with a +1 in Liu and a -1 in MPQA, that are matched by MPQA, are: solicitude, flair, funny, resurgent, untouched, tenderness, giddy, vulnerable, joke, shimmer, spurn, craven, aweful, backwoods, backwood, back-woods, back-wood, back-logged, backaches, backache, backaching, backbite, tingled, glower, and gainsay.

In Panel II of \ref{fig:main}, the words with a +1 in Liu and a -1 in LIWC, that are matched by LIWC, are: stunned, fiery, defeated, defeating, defeats, defeat, doubtless, dominated, dominates, dominate, dumbfounded, dumbfounding, faultless, foolproof, problem-free, problem-solver, risk-free, blameless, envy, trivially, trouble-free, tougher, toughest, tough, low-priced, low-price, low-risk, low-cost, lower-priced, geekier, geeky, guiltless, obsessions, and obsession.
The words with a -1 in Liu and a +1 in LIWC, that are matched by LIWC, are: silly, madly, sillily, dearth, challenging, cheerless, faithless, flirty, flirt, funnily, funny, tenderness, laughable, laughably, laughingstock, grating, opportunistic, joker, and joke.

In the off-diagonal bins for all of the $\pm 1$ dictionaries, we see many of the same words.
Again MPQA stem matches are disparagingly broad.
We also find matches by LIWC that are concerning, and should in all likelihood be removed from the dictionary.

\clearpage
\pagebreak

\section*{S3 Appendix: Coverage for all corpuses} \label{supp:coverage}
We provide coverage plots for the other three corpuses.

\begin{figure*}[!htb]
  \includegraphics[width=0.98\textwidth]{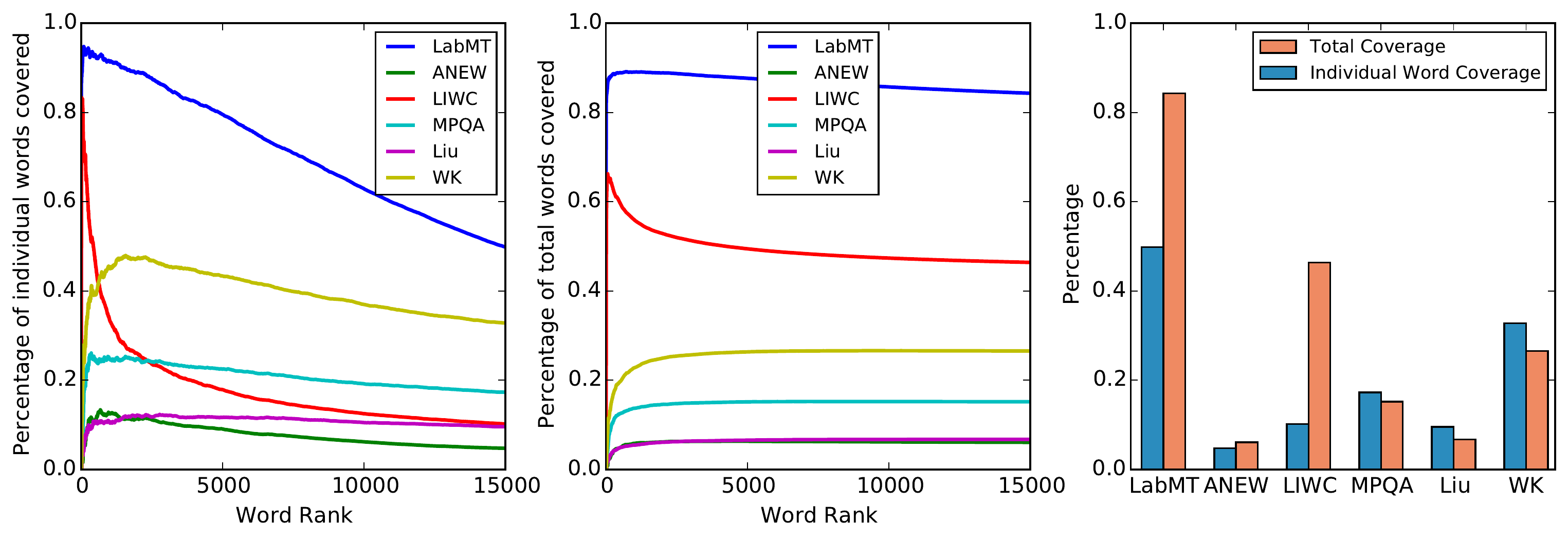}
    \caption[]{
      Coverage of the words on twitter by each of the dictionaries.
  }
  \label{fig:coverage_twitter}
\end{figure*}

\begin{figure*}[!htb]
  \includegraphics[width=0.96\textwidth]{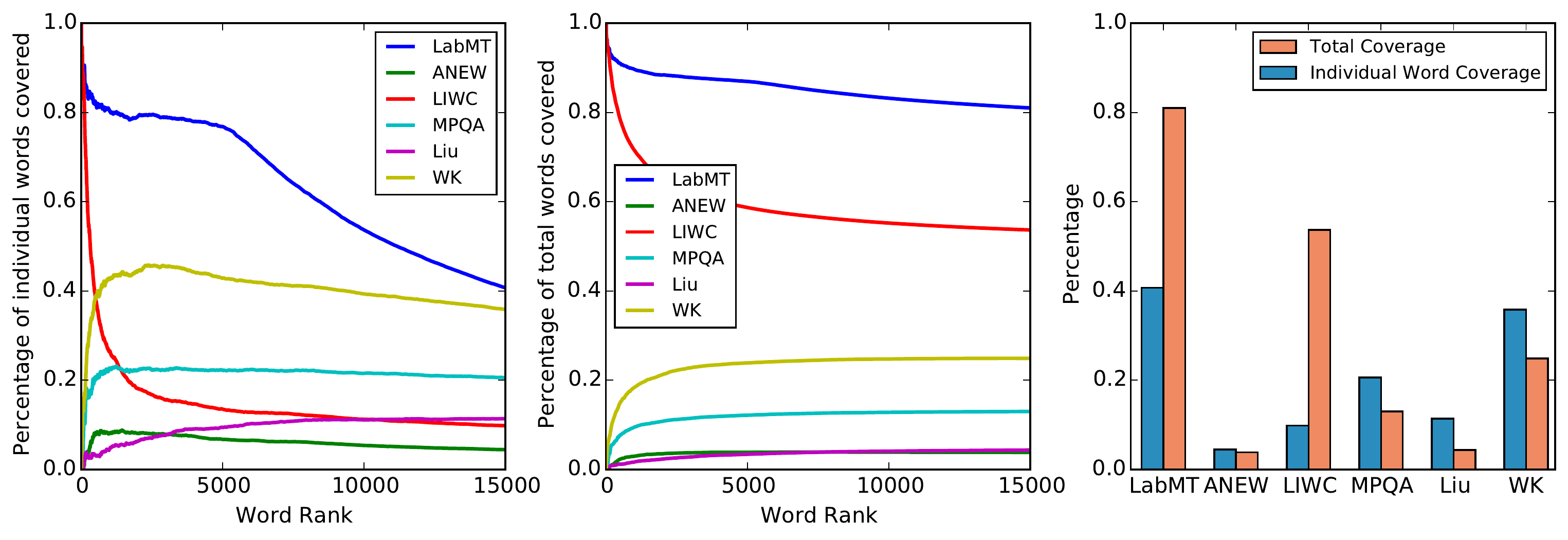}
    \caption[]{
    Coverage of the words in Google books by each of the dictionaries.
  }
  \label{fig:coverage_gbooks}
\end{figure*}    

\begin{figure*}[!htb]
  \includegraphics[width=0.96\textwidth]{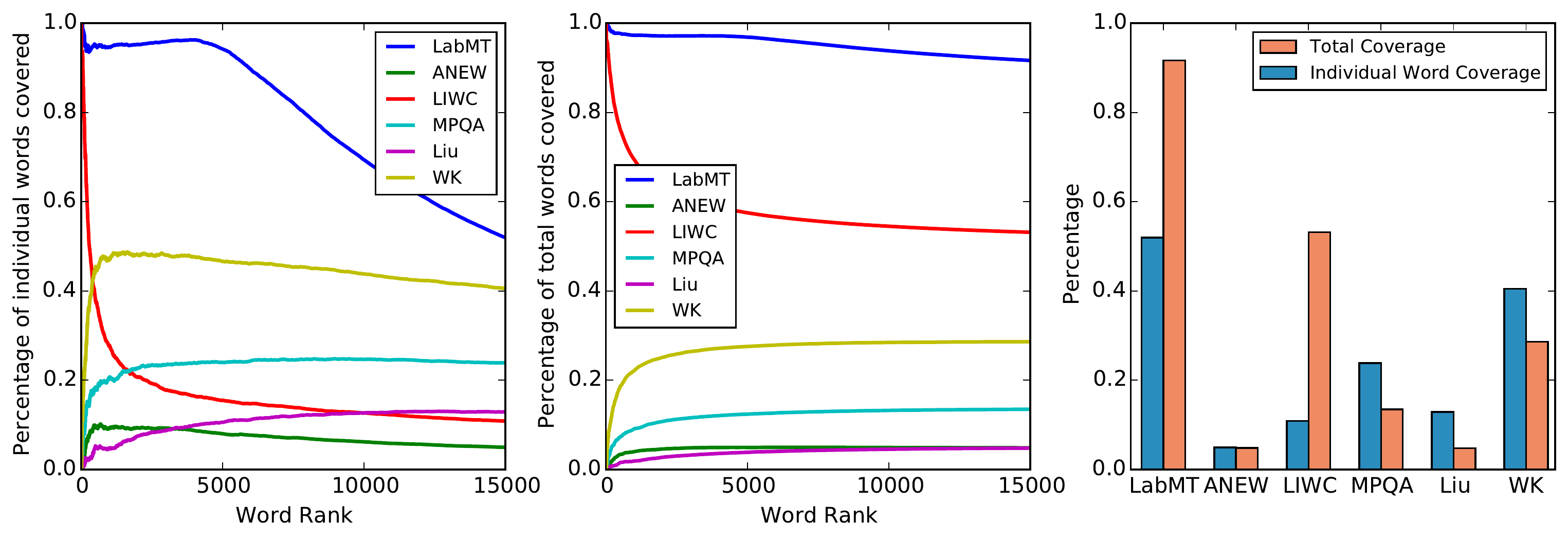}
    \caption[]{
    Coverage of the words in the New York Times by each of the dictionaries.
  }
  \label{fig:coverage_nyt}
\end{figure*}

\clearpage
\pagebreak

\section*{S4 Appendix: Sorted New York Times rankings} \label{supp:nyt}

\begin{figure*}[!htb]
  \includegraphics[width=0.98\textwidth]{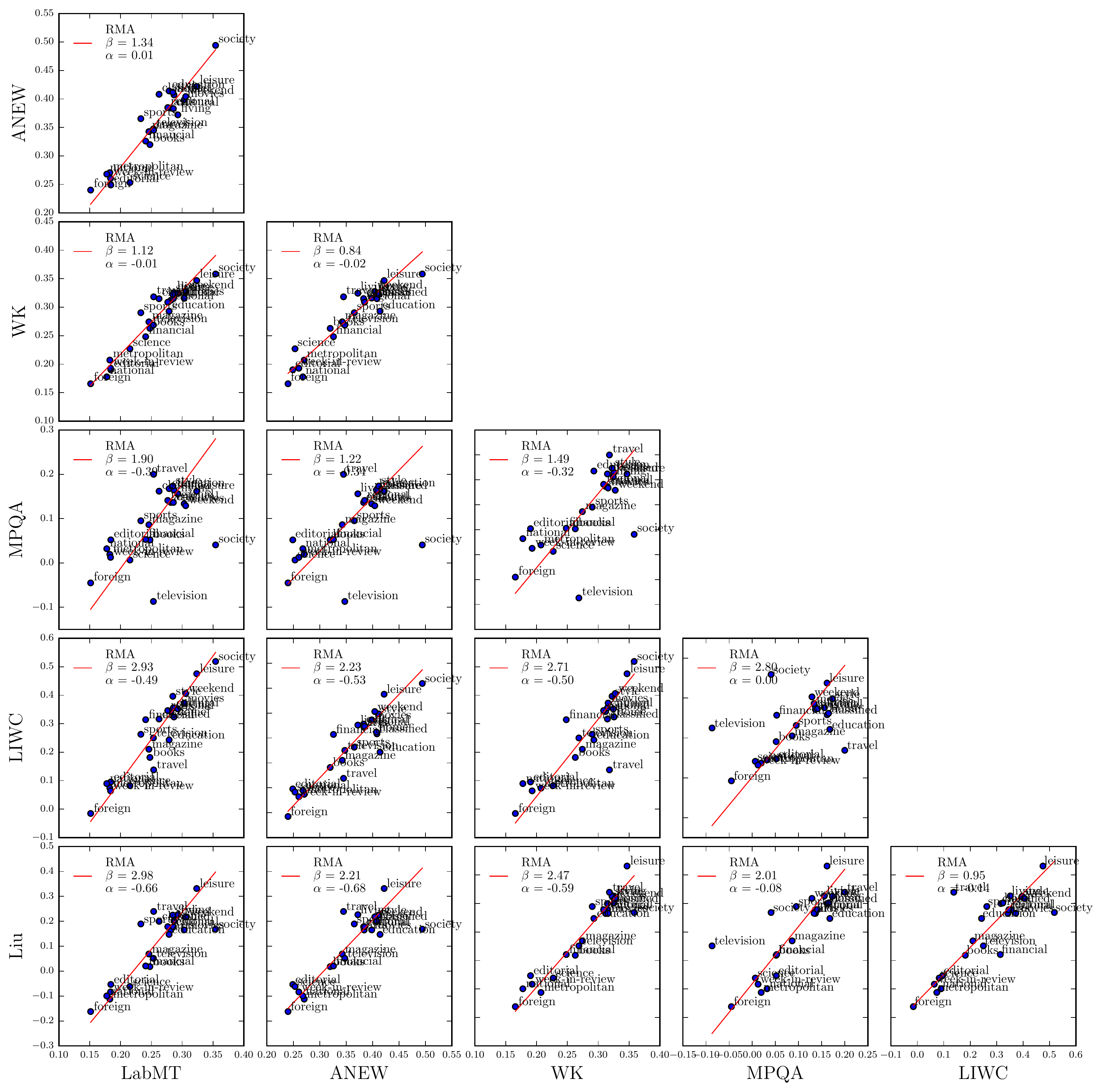}
    \caption[]{
      NYT Sections scatterplot.
      The RMA fit $\alpha$ and $\beta$ for the formula $y = \alpha + \beta x$.
      For the sake of comparison, we normalized each dictionary to the range [-.5,.5] by subtracting the mean score (5 or 0) and dividing by the range (8 or 2).
  }
  \label{fig:nyt_scatter_all}
\end{figure*}

\begin{figure*}[!htb]
  \includegraphics[width=0.96\textwidth]{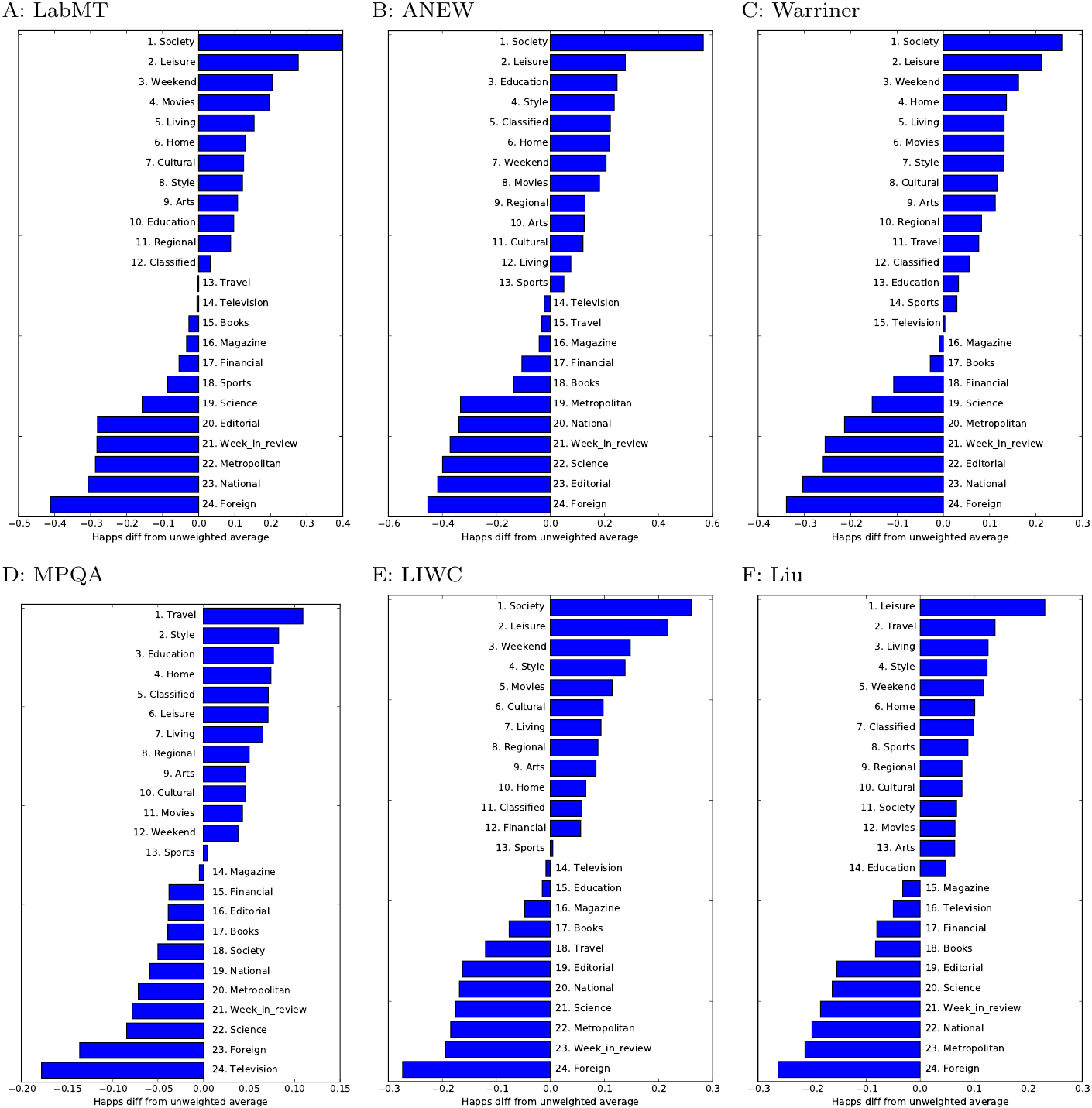}
    \caption[]{
    Sorted bar charts ranking each of the 24 New York Times Sections for each dictionary tested.
  }
  \label{fig:nyt_barcharts_all}
\end{figure*}

\clearpage
\pagebreak

\section*{S5 Appendix: Movie Review Distributions} \label{supp:movies}

\begin{figure*}[!htb]
  \centering
  \includegraphics[width=0.96\textwidth]{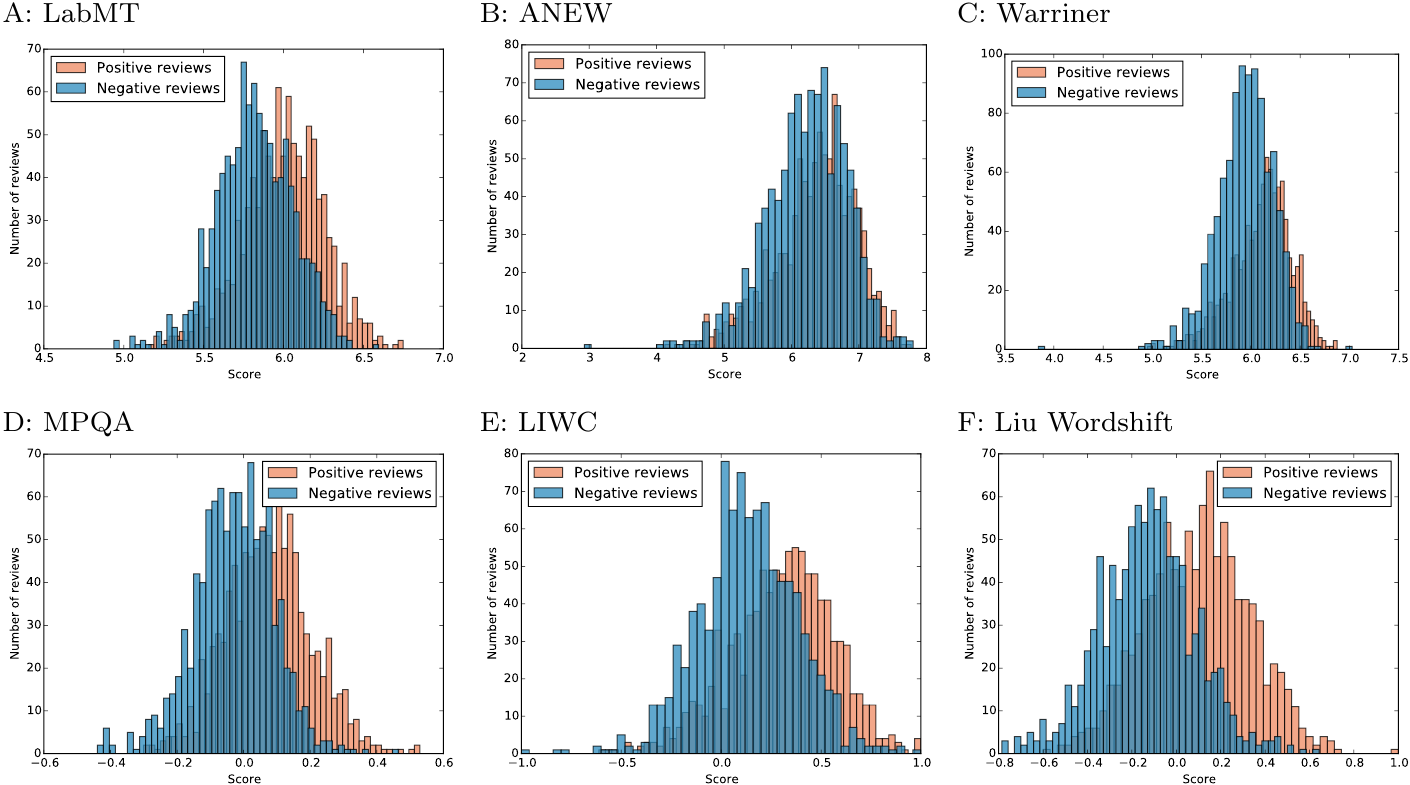}  
  \caption[]{
    Binned scores for each review by each corpus with a stop value of $\Delta _h = 1.0$.
  }
  \label{fig:moviereview-dist-1}
\end{figure*}

Here we examine the distributions of movie review scores.
These distributions are each summarized by their mean and standard deviation in panels of Figure 2 for each dictionary.
For example, the left most error bar of each panel in Figure 2 shows the standard deviation around the mean for the distribution of individual review scores (Figure \ref{fig:moviereview-dist-1}).

\begin{figure*}[!htb]
  \centering
  \includegraphics[width=0.96\textwidth]{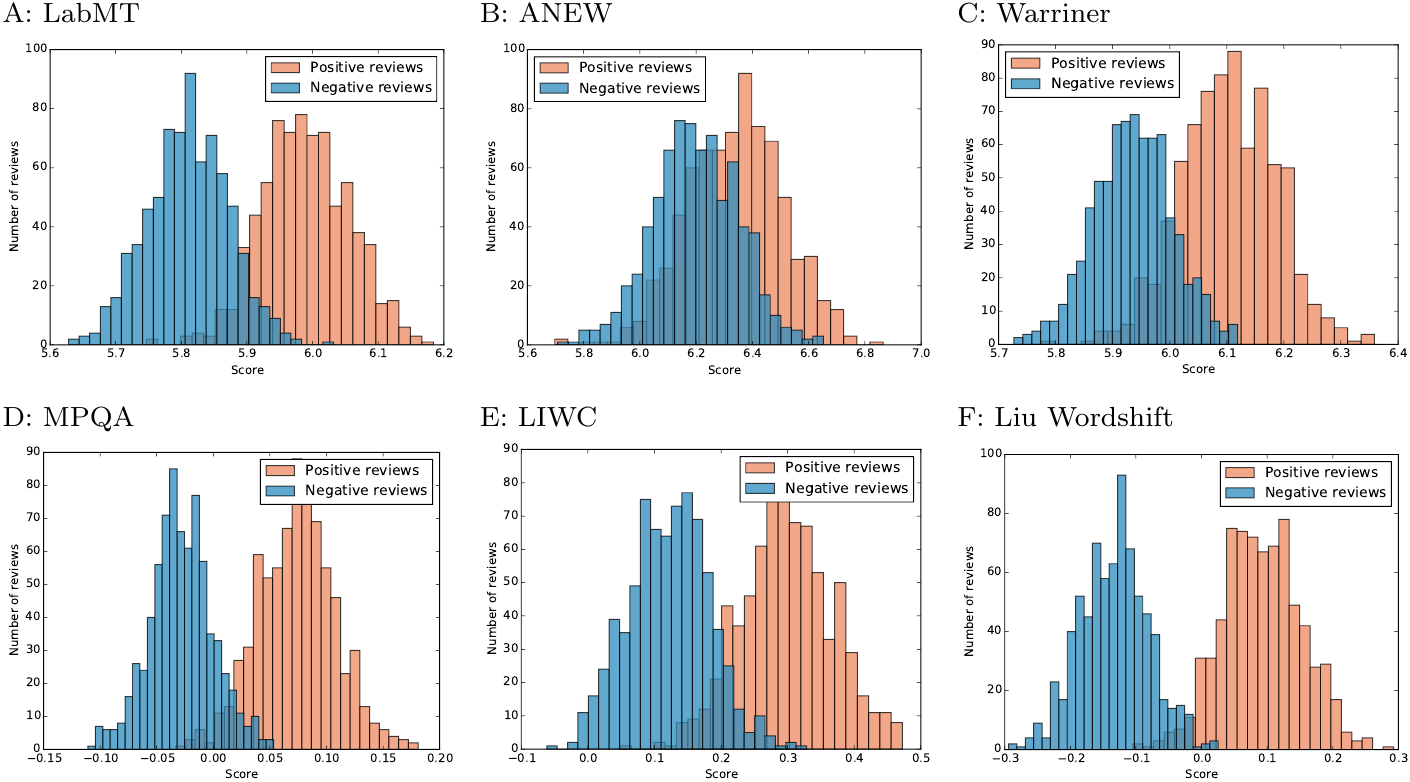}
  \caption[]{
    Binned scores for samples of 15 concatenated random reviews.
    Each dictionary uses stop value of $\Delta _h = 1.0$.
  }
  \label{fig:moviereview-dist-15}
\end{figure*}

\begin{figure*}[!htb]
  \centering
  \includegraphics[width=0.45\textwidth]{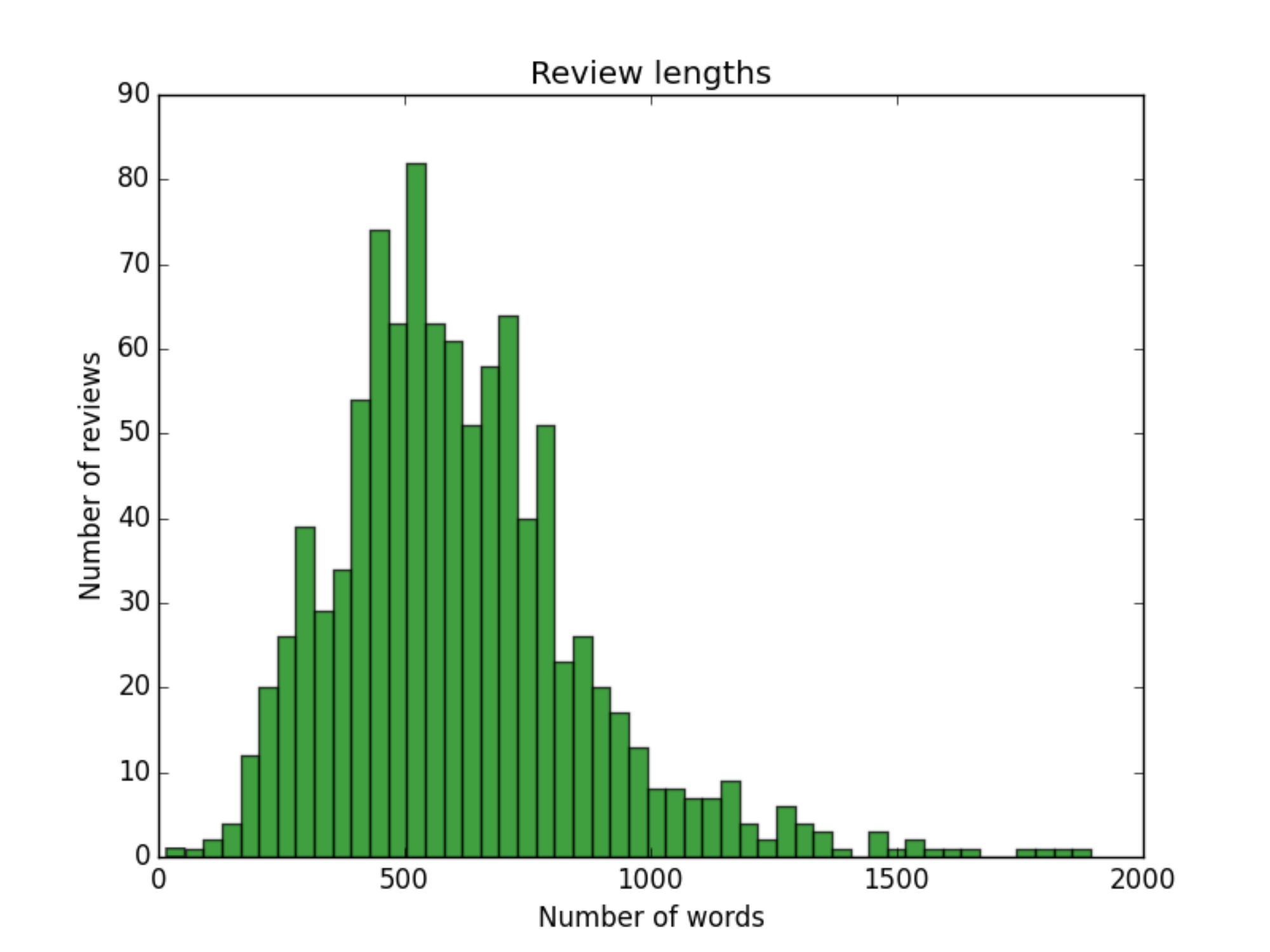}
  \caption[]{
    Binned length of positive reviews, in words.
  }
  \label{fig:moviereview-lengths}
\end{figure*}

\clearpage
\pagebreak

\section*{S6 Appendix: Google Books correlations and word shifts} \label{supp:gbooks}

\begin{figure*}[!htb]
  \includegraphics[width=0.48\textwidth]{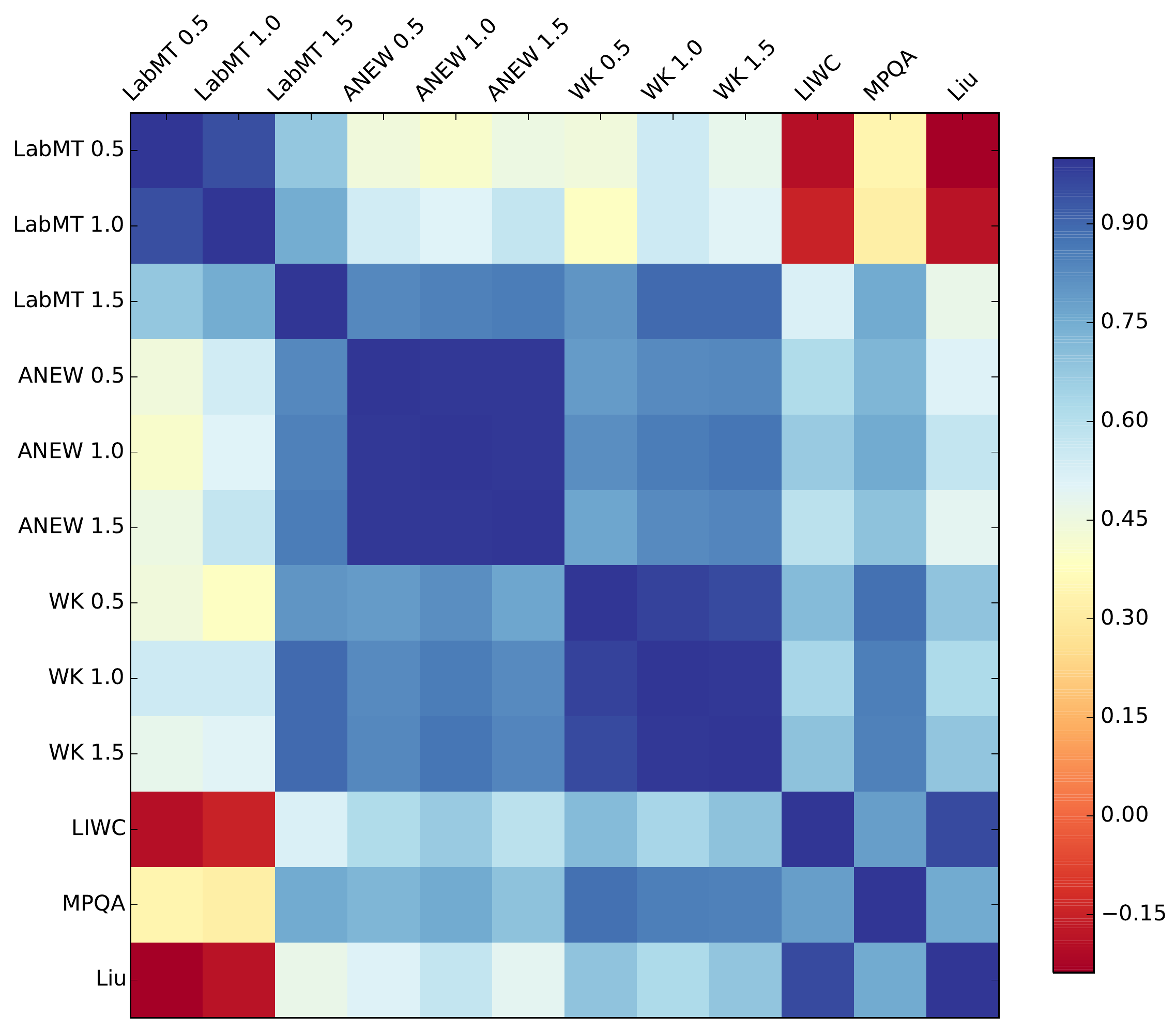}
    \caption[]{
    Google Books correlations.
    Here we include correlations for the google books time series, and word shifts for selected decades (1920's,1940's,1990's,2000's).

  }
  \label{fig:gbooks_correlations}
\end{figure*}

\begin{figure*}[!htb]
 \includegraphics[width=0.98\textwidth]{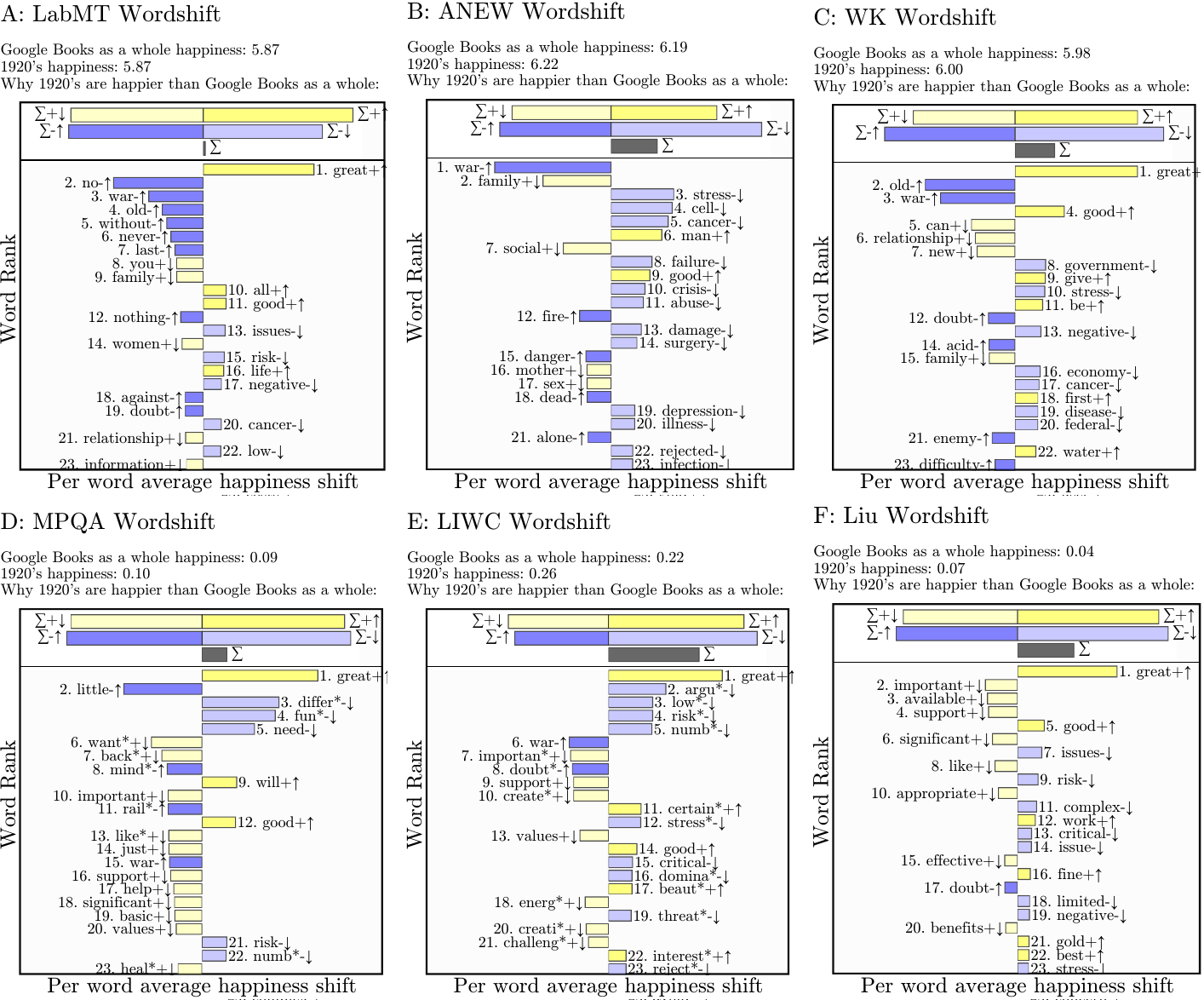}
  \caption[]{
    Google Books shifts in the 1920's against the baseline of Google Books.
  }
  \label{fig:gbooks-shifts-1920}
\end{figure*}

\begin{figure*}[!htb]
 \includegraphics[width=0.98\textwidth]{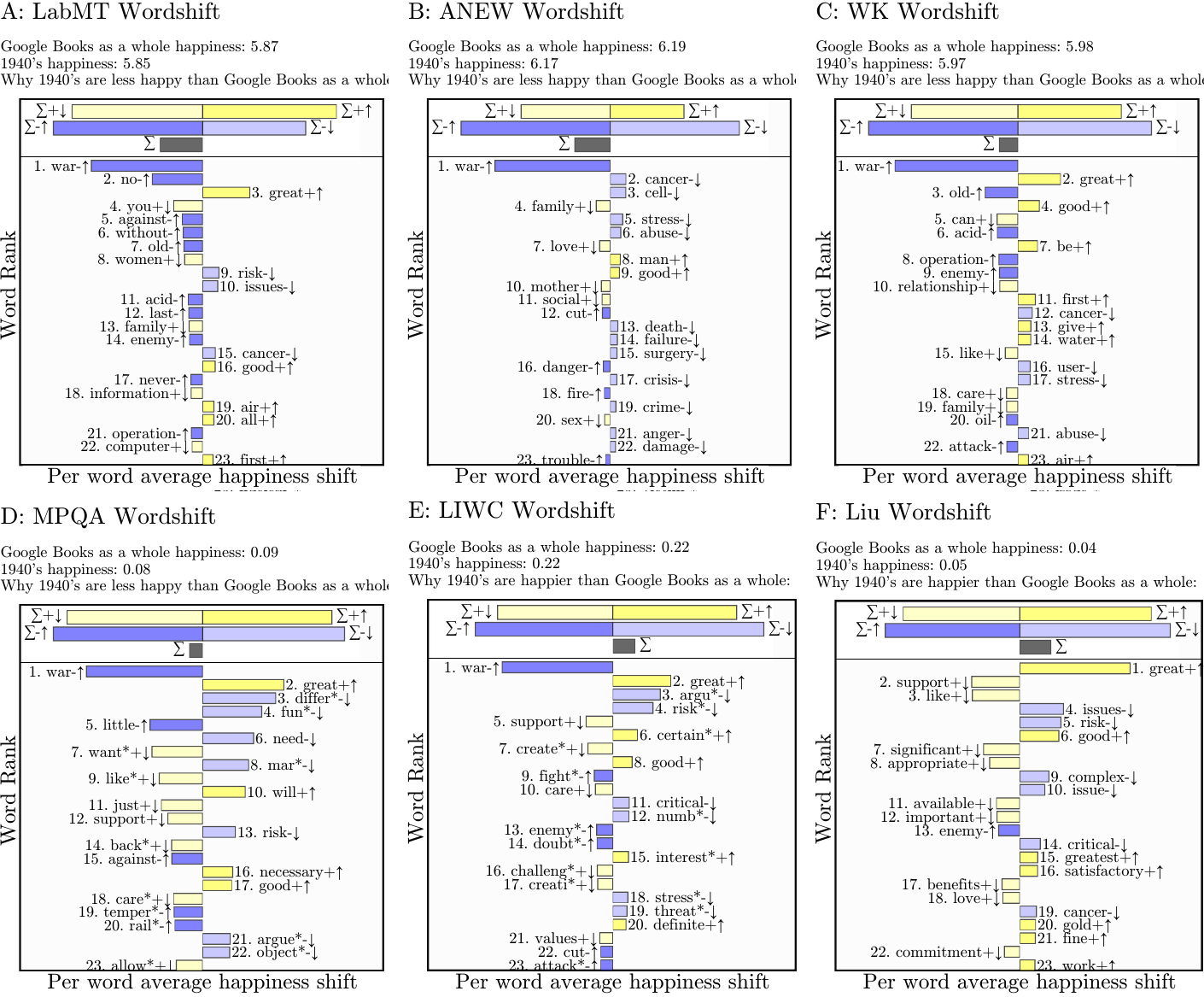}
  \caption[]{
    Google Books shifts in the 1940's against the baseline of Google Books.
  }
  \label{fig:gbooks-shifts-1940}
\end{figure*}

\begin{figure*}[!htb]
 \includegraphics[width=0.98\textwidth]{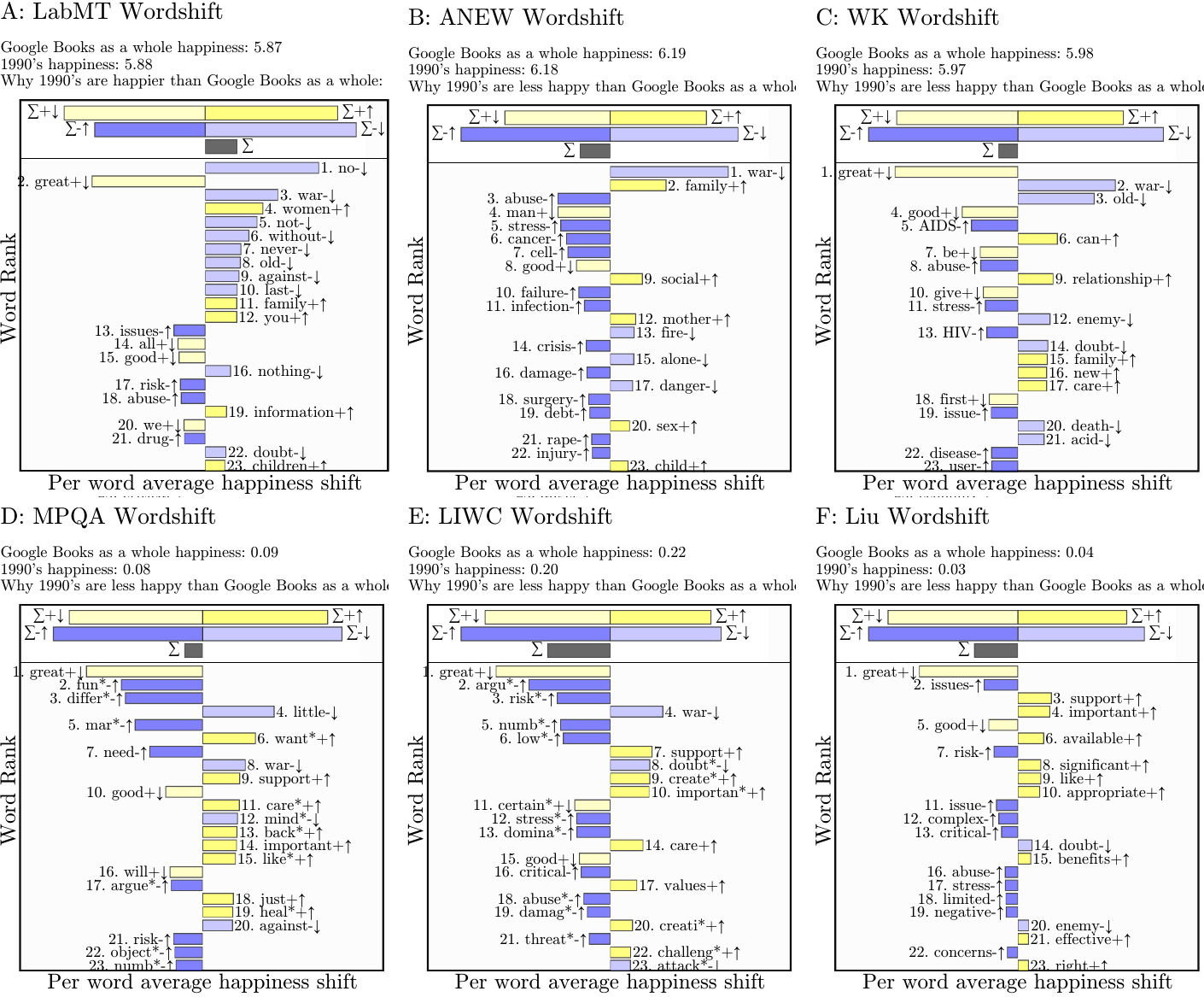}
  \caption[]{
    Google Books shifts in the 1990's against the baseline of Google Books.
  }
  \label{fig:gbooks-shifts-1990}
\end{figure*}

\begin{figure*}[!htb]
 \includegraphics[width=0.98\textwidth]{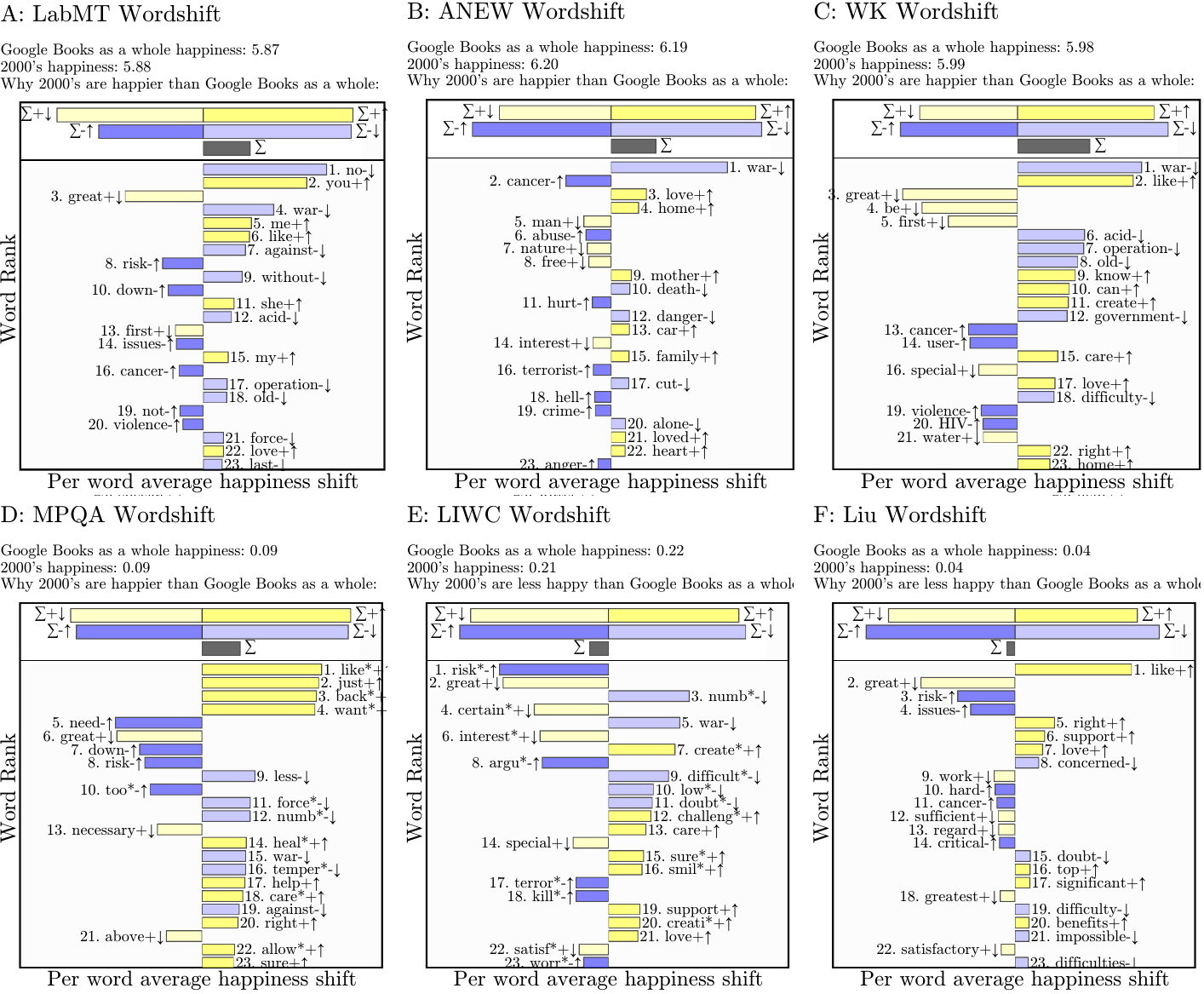}
  \caption[]{
    Google Books shifts in the 2000's against the baseline of Google Books.
  }
  \label{fig:gbooks-shifts-2000}
\end{figure*}

\clearpage
\pagebreak

\section*{S7 Appendix: Additional Twitter time series, correlations, and shifts} \label{supp:twitter}

First, we present additional Twitter time series:

\begin{figure*}[!htb]
  \centering
  \includegraphics[width=0.98\textwidth]{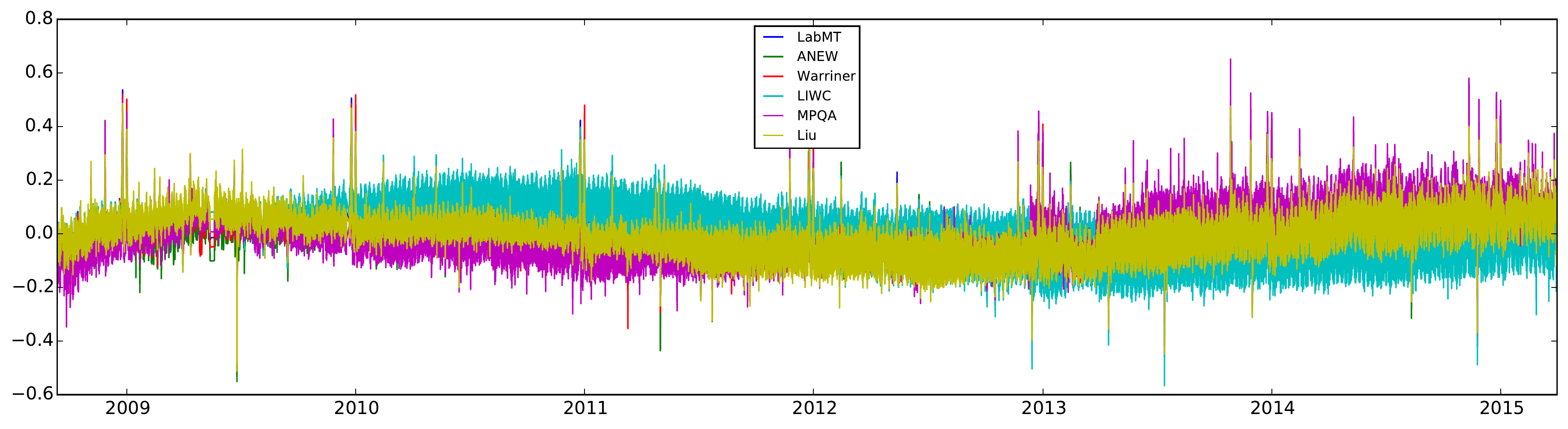}
  \caption[]{
    Normalized time series on Twitter using $\Delta _h$ of 1.0 for all.
    For resolution of 3 hours.
    We do not include any of the time series with resolution below 3 hours here because there are too many data points to see.
  }
  \label{fig:twitter_timeseries_2}
\end{figure*}

\begin{figure*}[!htb]
  \centering
  \includegraphics[width=0.98\textwidth]{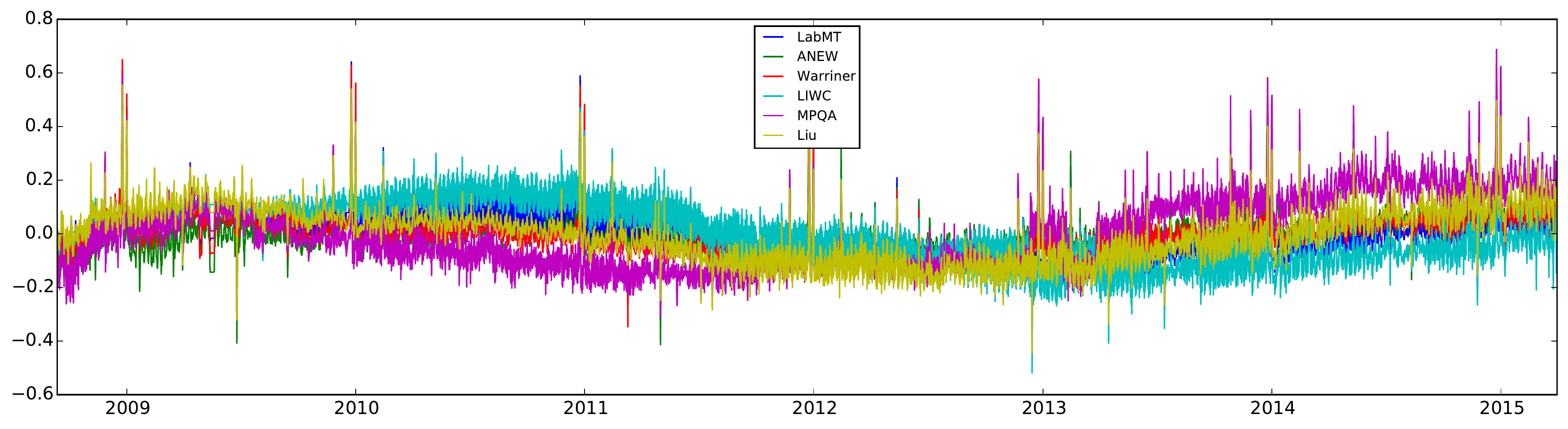}
  \caption[]{
    Normalized time series on Twitter using $\Delta _h$ of 1.0 for all.
    For resolution of 12 hours.
  }
  \label{fig:twitter_timeseries_3}
\end{figure*}

Next, we take a look at more correlations:

\begin{figure*}[!htb]
  \includegraphics[width=0.96\textwidth]{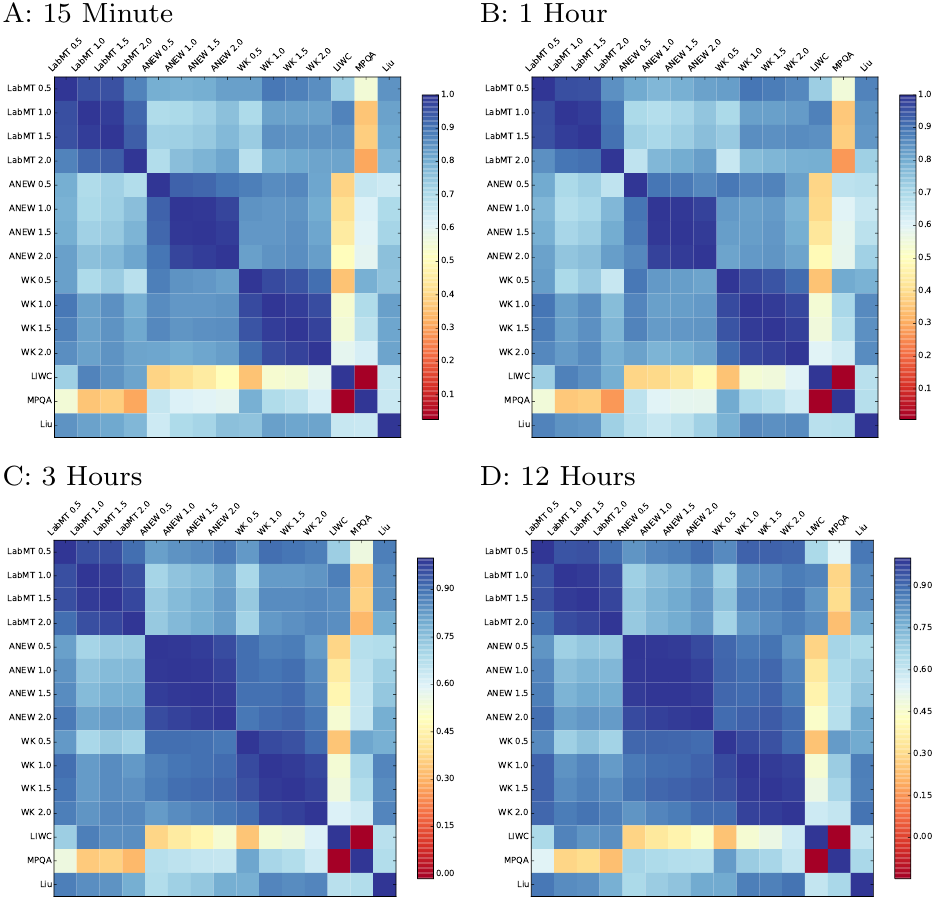}
    \caption[]{
    Pearson's $r$ correlation between Twitter time series for all resolutions below 1 day.
  }
  \label{fig:twitter_correlation_others}
\end{figure*}

Now we include word shift graphs that are absent from the manuscript itself.

\begin{figure*}[!htb]
  \centering
  \includegraphics[width=0.98\textwidth]{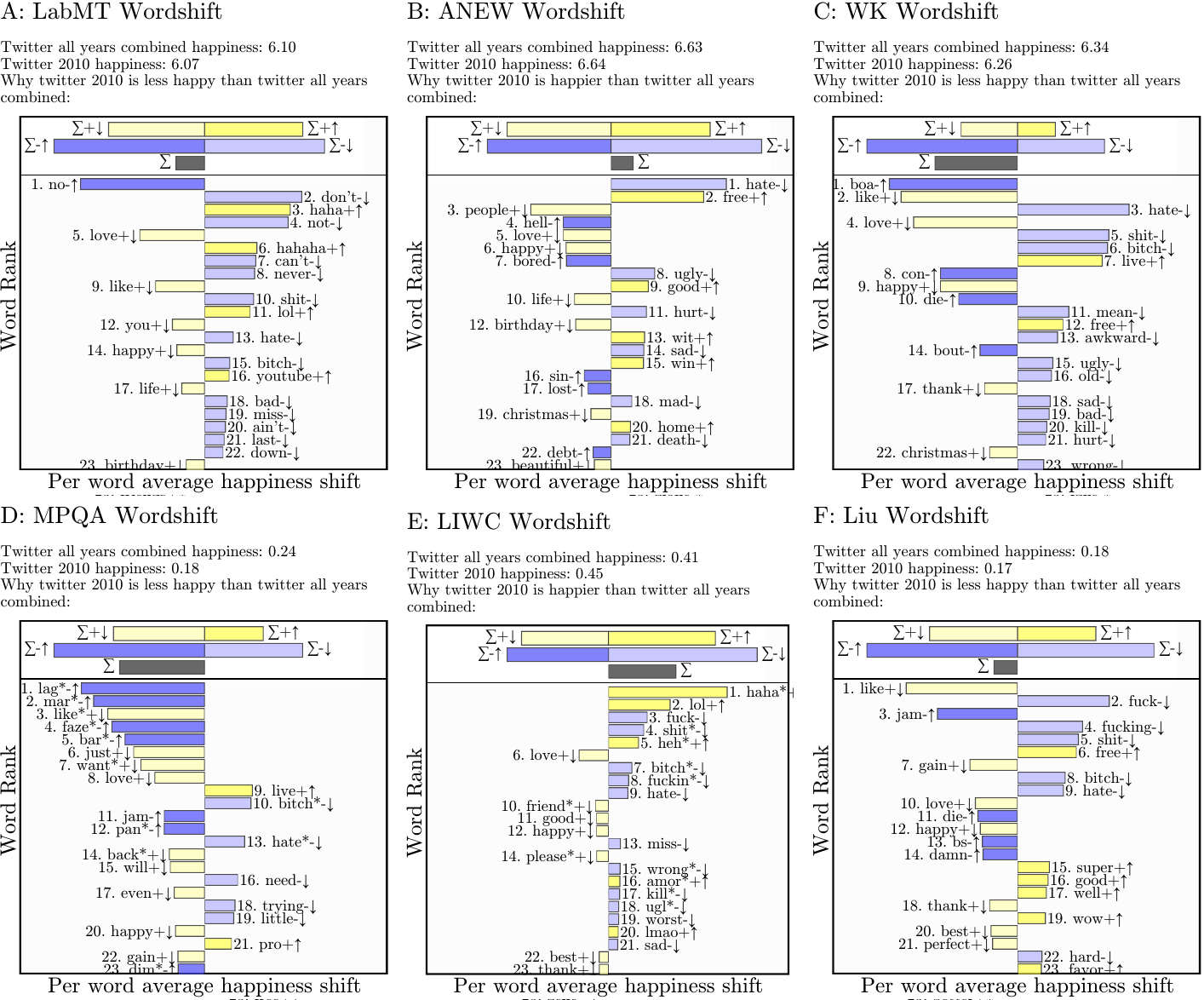}
  \caption[]{
    Word Shifts for Twitter in 2010.
    The reference word usage is all of Twitter (the 10\% Gardenhose feed) from September 2008 through April 2015, with the word usage normalized by year.
  }
  \label{fig:twitter-shift-2010}
\end{figure*}

\begin{figure*}[!htb]
  \centering
  \includegraphics[width=0.98\textwidth]{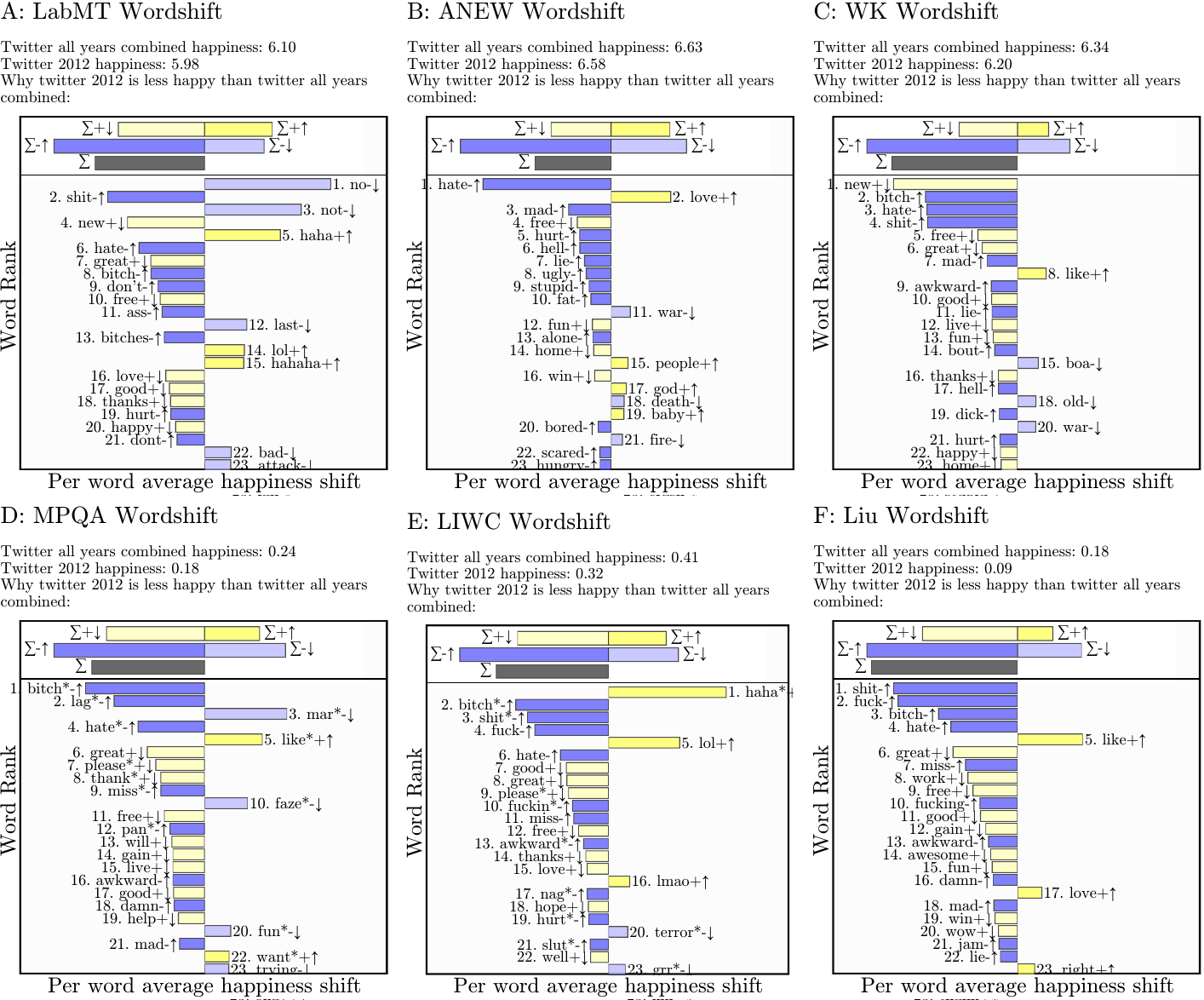}
  \caption[]{
    Word Shifts for Twitter in 2012.
    The reference word usage is all of Twitter (the 10\% Gardenhose feed) from September 2008 through April 2015, with the word usage normalized by year.
  }
  \label{fig:twitter-shift-2012}
\end{figure*}

\begin{figure*}[!htb]
  \centering
  \includegraphics[width=0.98\textwidth]{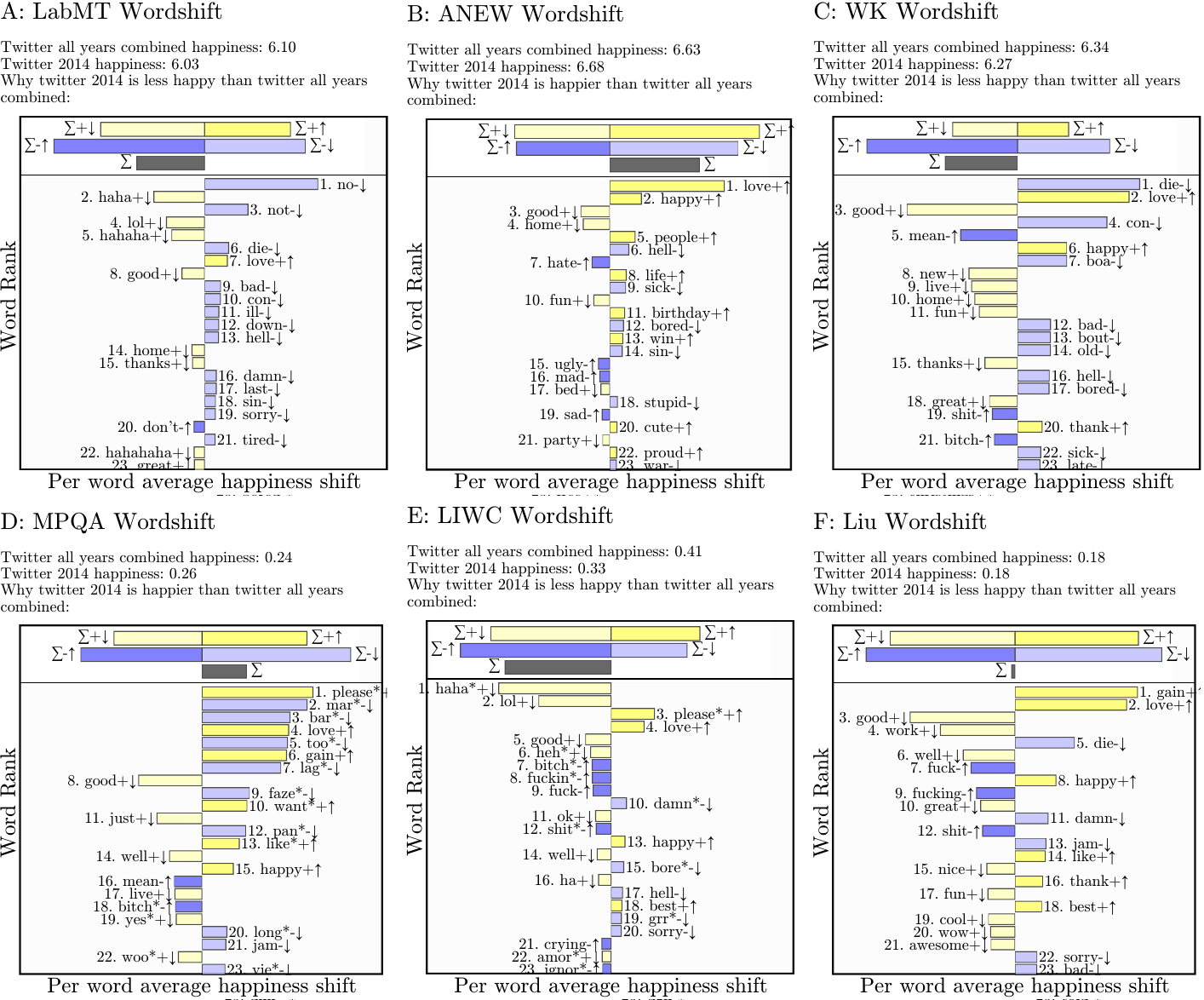}
  \caption[]{
    Word Shifts for Twitter in 2014.
    The reference word usage is all of Twitter (the 10\% Gardenhose feed) from September 2008 through April 2015, with the word usage normalized by year.
  }
  \label{fig:twitter-shift-2014}
\end{figure*}

Finally, we include the results of each dictionary applied to a set of annotated Twitter data.
We apply sentiment dictionaries to rate individual tweets and classify a tweet as positive (negative) if the tweet rating is greater (less) than the average of all scores in dictionary.

\begin{table*}[!htb]
  \begin{adjustwidth}{\pnastableadjust in}{0in}
    \centering
\begin{tabular}{l | l | c | c | c | c}
Rank & Dictionary & \% tweets scored & F1 of tweets scored & Calibrated F1 & Overall F1\\
\hline
1. & Sent140Lex & 100.0 & 0.89 & 0.88 & 0.89\\
2. & labMT & 100.0 & 0.69 & 0.78 & 0.69\\
3. & HashtagSent & 100.0 & 0.67 & 0.64 & 0.67\\
4. & SentiWordNet & 98.6 & 0.67 & 0.68 & 0.67\\
5. & VADER & 81.3 & 0.75 & 0.81 & 0.61\\
6. & SentiStrength & 73.9 & 0.83 & 0.81 & 0.61\\
7. & SenticNet & 97.3 & 0.61 & 0.64 & 0.59\\
8. & Umigon & 67.1 & 0.87 & 0.85 & 0.58\\
9. & SOCAL & 82.2 & 0.71 & 0.75 & 0.58\\
10. & WDAL & 99.9 & 0.58 & 0.64 & 0.58\\
11. & AFINN & 73.6 & 0.78 & 0.80 & 0.57\\
12. & OL & 66.7 & 0.83 & 0.82 & 0.55\\
13. & MaxDiff & 94.1 & 0.58 & 0.70 & 0.54\\
14. & EmoSenticNet & 96.0 & 0.56 & 0.59 & 0.54\\
15. & MPQA & 73.2 & 0.73 & 0.72 & 0.53\\
16. & WK & 96.5 & 0.53 & 0.72 & 0.51\\
17. & LIWC15 & 61.8 & 0.81 & 0.78 & 0.50\\
18. & Pattern & 69.0 & 0.71 & 0.75 & 0.49\\
19. & GI & 67.6 & 0.72 & 0.70 & 0.49\\
20. & LIWC07 & 60.3 & 0.80 & 0.75 & 0.48\\
21. & LIWC01 & 54.3 & 0.83 & 0.75 & 0.45\\
22. & EmoLex & 59.4 & 0.73 & 0.69 & 0.43\\
23. & ANEW & 64.1 & 0.65 & 0.68 & 0.42\\
24. & USent & 4.5 & 0.74 & 0.73 & 0.03\\
25. & PANAS-X & 1.7 & 0.88 & -- & 0.01\\
26. & Emoticons & 1.4 & 0.72 & 0.77 & 0.01\\
\end{tabular}
    \end{adjustwidth}
  \caption{Ranked results of sentiment dictionary performance on individual tweets from STS-Gold dataset (Saif, 2013).
    We report the percentage of tweets for which each dictionary contains at least 1 entry, the F1 score on those tweets, and the overall classification F1 score.
    The calibrated F1 score tunes the decision threshold between positive and negative tweets with a random 10\% training sample.}
\label{tbl:STS}
\end{table*}

\clearpage
\pagebreak

\section*{S8 Appendix: Naive Bayes results and derivation} \label{supp:bayes}

We now provide more details on the implementation of Naive Bayes, a derivation of the linearity structure, and more results from the classification of Movie Reviews.

First, to implement a binary Naive Bayes classifier for a collection of documents, we denote each of the $N$ words in the given document $T$ as $w_i$, the word frequency as $f_i(T)$, and class labels $c_1,c_2$.
The probability of a document $T$ belonging to class $c_1$ can be written as
$$ P(c_1 | T) = \frac{P(c_1)P(T|c_1)}{P(T)}. $$
Since we do not know $P(T|c_1)$ explicitly, we make the \textit{naive} assumption that each word appears independently, and thus write
$$ P(c_1 | T) = \frac{P(c_1)\cdot \left [ P(f_1(T)|c_1) \cdot P(f_2(T)|c_1) \cdots P(f_N(T)|c_1)\right] }{P(T)}. $$
Since we are only interested in comparing $P(c_1 | T)$ and $P(c_2 | T)$, we disregard the shared denominator and have
$$ P(c_1 | T) \propto P(c_1)\cdot \left [ P(f_1(T)|c_1) \cdot P(f_2(T)|c_1) \cdots P(f_N(T)|c_1)\right] . $$
Finally we say that document $T$ belongs to class $c_1$ if $P(c_1 | T) > P(c_2 | T)$.
Given that the probabilities of individual words are small, to avoid machine truncation error we compute these probabilities in log space, such that the product of individual word likelihoods becomes a sum
$$ \log P(c_1 | T) \propto \log P(c_1) + \sum_{i=1}^N P(f_i(T)|c_1)  . $$
Assigning a classification of class $c_1$ if $P(c_1 | T) > P(c_2 | T)$ is the same as saying that the difference between the two is positive, i.e. $P(c_1 | T) - P(c_2 | T) > 0$ and since the logarithm is monotonic, $\log P(c_1 | T) - \log P(c_2 | T) > 0$.
To examine how individual words contribute to this difference, we can write
\begin{align*} 0 &< \log P(c_1 | T) - \log P(c_2 | T) \\
  &\propto \log P(c_1) + \sum_{i=1}^N \log P(f_i(T)|c_1) -\log P(c_2) - \sum_{i=1}^N \log P(f_i(T)|c_2)\\
  &\propto \log P(c_1) -\log P(c_2) + \sum_{i=1}^N \left [\log P(f_i(T)|c_1) \log P(f_i(T)|c_2) \right ]\\
  &\propto \log \frac{P(c_1)}{P(c_2)} + \sum_{i=1}^N \log \frac{P(f_i(T)|c_1)}{P(f_i(T)|c_2)}.\end{align*}
We can see from the above that the contribution of each word $w_i$ (or more accurately, the likelihood of frequency in document $T$ being of class $c$ as $P(f_i(T)|c_1)$) is a linear constituent of the classification.

Next, we include the detailed results of the Naive Bayes classifier on the Movie Review corpus.

\begin{figure*}[!htb]
  \includegraphics[width=0.96\textwidth]{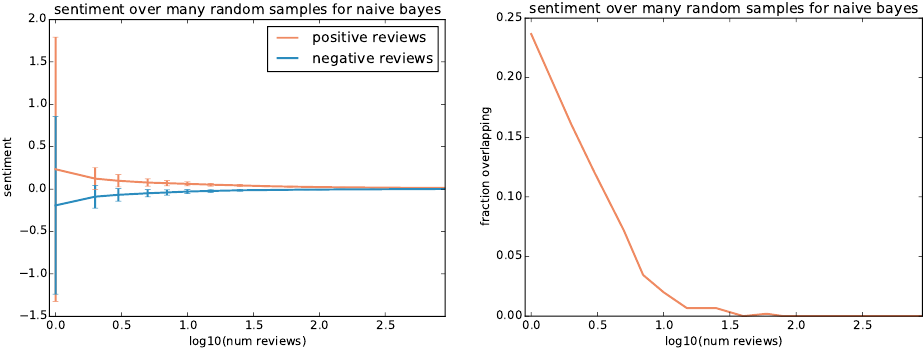}
    \caption[]{
    Results of the NB classifier on the Movie Reviews corpus.
  }
  \label{fig:NBresult}
\end{figure*}

\begin{figure*}[!htb]
  \includegraphics[width=0.96\textwidth]{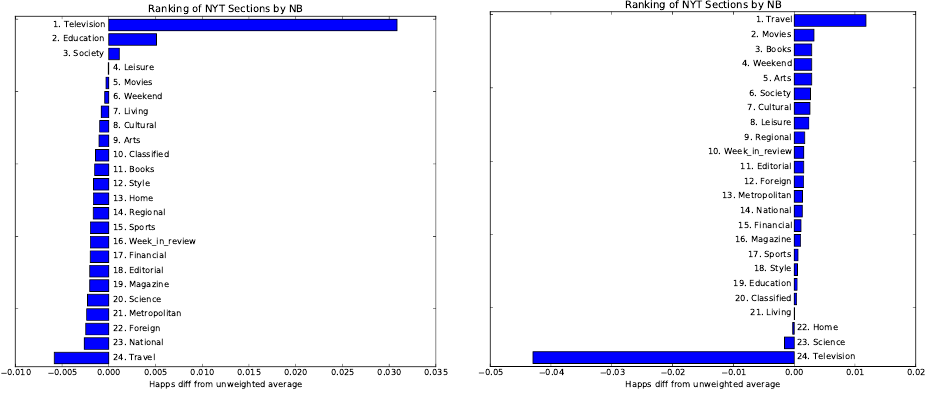}
    \caption[]{
    NYT Sections ranked by Naive Bayes in two of the five trials.
  }
  \label{fig:NBbars}
\end{figure*}

\begin{table*}[!htb]
  \begin{adjustwidth}{\pnastableadjust in}{0in}
    \centering
    \begin{tabular}{ | p{2cm} p{2cm} | p{2cm} p{2cm} |  }
  \hline
  \multicolumn{4}{|c|}{Most informative}  \\
  \hline
  \multicolumn{2}{|c}{Positive} & \multicolumn{2}{|c}{Negative} \\
  \hline
  Word & Value & Word & Value  \\
  \hline
  27.27 & flynt & 20.21 & godzilla  \\
26.33 & truman & 15.95 & werewolf \\
20.68 & charles & 13.83 & gorilla \\
15.04 & event & 13.83 & spice \\
14.10 & shrek & 13.83 & memphis \\
13.16 & cusack & 13.83 & sgt \\
13.16 & bulworth & 12.76 & jennifer \\
13.16 & robocop & 12.76 & hill \\
12.22 & jedi & 11.70 & max \\
12.22 & gangster & 11.70 & 200 \\
\hline
\end{tabular}
\\[3ex]
\begin{tabular}{ | p{2cm} p{2cm} | p{2cm} p{2cm} |  }
\hline
  \multicolumn{4}{|c|}{NYT Society} \\
  \hline
  \multicolumn{2}{|c}{Positive} & \multicolumn{2}{|c}{Negative}  \\
  \hline  
  Word & Value & Word & Value \\
  \hline
  26.08 & truman & 20.40 & godzilla \\
  20.49 & charles & 12.88 & hill  \\
  12.11 & gangster & 12.88 & jennifer  \\
  10.25 & speech & 10.73 & fatal  \\
  9.32 & melvin & 8.59 & freddie  \\
  8.85 & wars & 8.59 & = \\
  7.45 & agents & 8.59 & mess \\
  6.52 & dance & 8.59 & gene \\
  6.52 & bleak & 8.59 & apparent \\
  6.52 & pitt & 7.51 & travolta \\
  \hline
\end{tabular}
    \end{adjustwidth}
  \caption{Trial 1 of Naive Bayes trained on a random 10\% of the movie review corpus, and applied to the New York Times Society section.
    We show the words which are used by the trained classifier to classify individual reviews (in corpus), and on the New York Times (out of corpus).
    In addition, we report a second trial in Table \ref{tbl:NB-2}, since Naive Bayes is trained on a random subset of data, to show the variation in individual words between trials (while performance is consistent).}
\label{tbl:NB-1}
\end{table*}

\begin{table*}[!htb]
  \begin{adjustwidth}{\pnastableadjust in}{0in}
    \centering
    \begin{tabular}{ | p{2cm} p{2cm} | p{2cm} p{2cm} |  }
  \hline
  \multicolumn{4}{|c|}{Most informative}  \\
  \hline
  \multicolumn{2}{|c}{Positive} & \multicolumn{2}{|c}{Negative} \\
  \hline
  Word & Value & Word & Value  \\
  \hline
  18.11 & shrek & 34.63 & west\\
  17.15 & poker & 24.14 & webb\\
  15.25 & shark & 18.89 & jackal\\
  14.29 & maggie & 17.84 & travolta\\
  13.34 & guido & 17.84 & woo\\
  13.34 & outstanding & 17.84 & coach\\
  13.34 & political & 16.79 & awful \\
  13.34 & journey & 16.79 & brenner \\
  13.34 & bulworth & 15.74 & gabriel \\
  12.39 & bacon & 15.74 & general's \\
\hline
\end{tabular}
\\[3ex]
\begin{tabular}{ | p{2cm} p{2cm} | p{2cm} p{2cm} |  }
\hline
  \multicolumn{4}{|c|}{NYT Society} \\
  \hline
  \multicolumn{2}{|c}{Positive} & \multicolumn{2}{|c}{Negative}  \\
  \hline  
  Word & Value & Word & Value \\
  \hline
  17.79 & poker & 33.39 & west \\
  13.84 & journey & 17.20 & coach \\
  13.84 & political & 17.20 & travolta \\
  8.90 & tribe & 15.18 & gabriel \\
  7.91 & tony & 12.14 & pointless \\
  7.91 & price & 9.44 & stupid \\
  7.91 & threat & 8.09 & screaming \\
  7.12 & titanic & 7.59 & mess \\
  6.92 & dicaprio & 7.42 & boring \\
  6.92 & kate & 7.08 & = \\
  \hline
\end{tabular}
    \end{adjustwidth}
  \caption{Trial 2 of Naive Bayes trained on a random 10\% of the movie review corpus, and applied to the New York Times Society section.
    We show the words which are used by the trained classifier to classify individual reviews (in corpus), and on the New York Times (out of corpus).
    This second trial is in addition to the first trial in Table \ref{tbl:NB-1}, since Naive Bayes is trained on a random subset of data, to show the variation in individual words between trials (while performance is consistent).}
\label{tbl:NB-2}
\end{table*}

\clearpage
\pagebreak

\section*{S9 Appendix: Movie review benchmark of additional dictionaries} \label{supp:additional}

Here, we present the accuracy of each dictionary applied to binary classification of Movie Reviews.

\begin{table*}[!htb]
  \begin{adjustwidth}{\pnastableadjust in}{0in}
      \centering
\begin{tabular}{l | l | c | c | c |}
Rank & Title & \% Scored & F1 Trained & F1 Untrained\\
\hline
1. & OL & 100 & 0.70 & 0.71\\
2. & HashtagSent & 100 & 0.67 & 0.66\\
3. & MPQA & 100 & 0.67 & 0.66\\
4. & SentiWordNet & 100 & 0.65 & 0.65\\
5. & labMT & 100 & 0.64 & 0.63\\
6. & AFINN & 100 & 0.67 & 0.63\\
7. & Umigon & 100 & 0.65 & 0.62\\
8. & GI & 100 & 0.65 & 0.61\\
9. & SOCAL & 100 & 0.71 & 0.60\\
10. & VADER & 100 & 0.67 & 0.60\\
11. & WDAL & 100 & 0.60 & 0.59\\
12. & SentiStrength & 100 & 0.63 & 0.58\\
13. & EmoLex & 100 & 0.65 & 0.56\\
14. & LIWC15 & 100 & 0.64 & 0.55\\
15. & LIWC01 & 100 & 0.65 & 0.54\\
16. & LIWC07 & 100 & 0.64 & 0.53\\
17. & Pattern & 100 & 0.73 & 0.52\\
18. & PANAS-X & 33 & 0.51 & 0.51\\
19. & Sent140Lex & 100 & 0.68 & 0.47\\
20. & SenticNet & 100 & 0.62 & 0.45\\
21. & ANEW & 100 & 0.57 & 0.36\\
22. & MaxDiff & 100 & 0.66 & 0.36\\
23. & EmoSenticNet & 100 & 0.58 & 0.34\\
24. & WK & 100 & 0.63 & 0.34\\
25. & Emoticons & 0 & -- & --\\
26. & USent & 40 & -- & --\\
\end{tabular}    \end{adjustwidth}
    \caption{Ranked performance of dictionaries on the Movie Review corpus.
    }
\label{tbl:MR-1}
\end{table*}

\begin{table*}[!htb]
    \begin{adjustwidth}{\pnastableadjust in}{0in}
      \centering
\begin{tabular}{l | l | c | c | c | c }
Rank & Title & \% Scored & F1 Trained of Scored & F1 Untrained of Scored & F1 Untrained, All\\
\hline
1. & HashtagSent & 100 & 0.55 & 0.55 & 0.55\\
2. & LIWC15 & 99 & 0.53 & 0.55 & 0.55\\
3. & LIWC07 & 99 & 0.53 & 0.55 & 0.54\\
4. & LIWC01 & 99 & 0.52 & 0.55 & 0.54\\
5. & labMT & 99 & 0.54 & 0.54 & 0.54\\
6. & Sent140Lex & 100 & 0.55 & 0.54 & 0.54\\
7. & SentiWordNet & 99 & 0.54 & 0.53 & 0.53\\
8. & WDAL & 99 & 0.53 & 0.53 & 0.52\\
9. & EmoLex & 95 & 0.54 & 0.55 & 0.52\\
10. & MPQA & 93 & 0.54 & 0.55 & 0.52\\
11. & SenticNet & 97 & 0.53 & 0.52 & 0.50\\
12. & SOCAL & 88 & 0.56 & 0.55 & 0.49\\
13. & EmoSenticNet & 98 & 0.52 & 0.46 & 0.45\\
14. & Pattern & 81 & 0.55 & 0.55 & 0.45\\
15. & GI & 80 & 0.55 & 0.55 & 0.44\\
16. & WK & 97 & 0.54 & 0.45 & 0.44\\
17. & OL & 76 & 0.56 & 0.57 & 0.44\\
18. & VADER & 79 & 0.56 & 0.55 & 0.43\\
19. & SentiStrength & 77 & 0.54 & 0.54 & 0.41\\
20. & MaxDiff & 83 & 0.54 & 0.49 & 0.41\\
21. & AFINN & 70 & 0.56 & 0.56 & 0.39\\
22. & ANEW & 63 & 0.52 & 0.48 & 0.30\\
23. & Umigon & 53 & 0.56 & 0.56 & 0.30\\
24. & PANAS-X & 1 & 0.53 & 0.53 & 0.01\\
25. & Emoticons & 0 & -- & -- & --\\
26. & USent & 2 & -- & -- & --\\
\end{tabular}    \end{adjustwidth}
    \caption{Ranked performance of dictionaries on the Movie Review corpus, broken into sentences.
    }
\label{tbl:MR-2}
\end{table*}

\begin{figure*}[!htb]
  \includegraphics[width=0.96\textwidth]{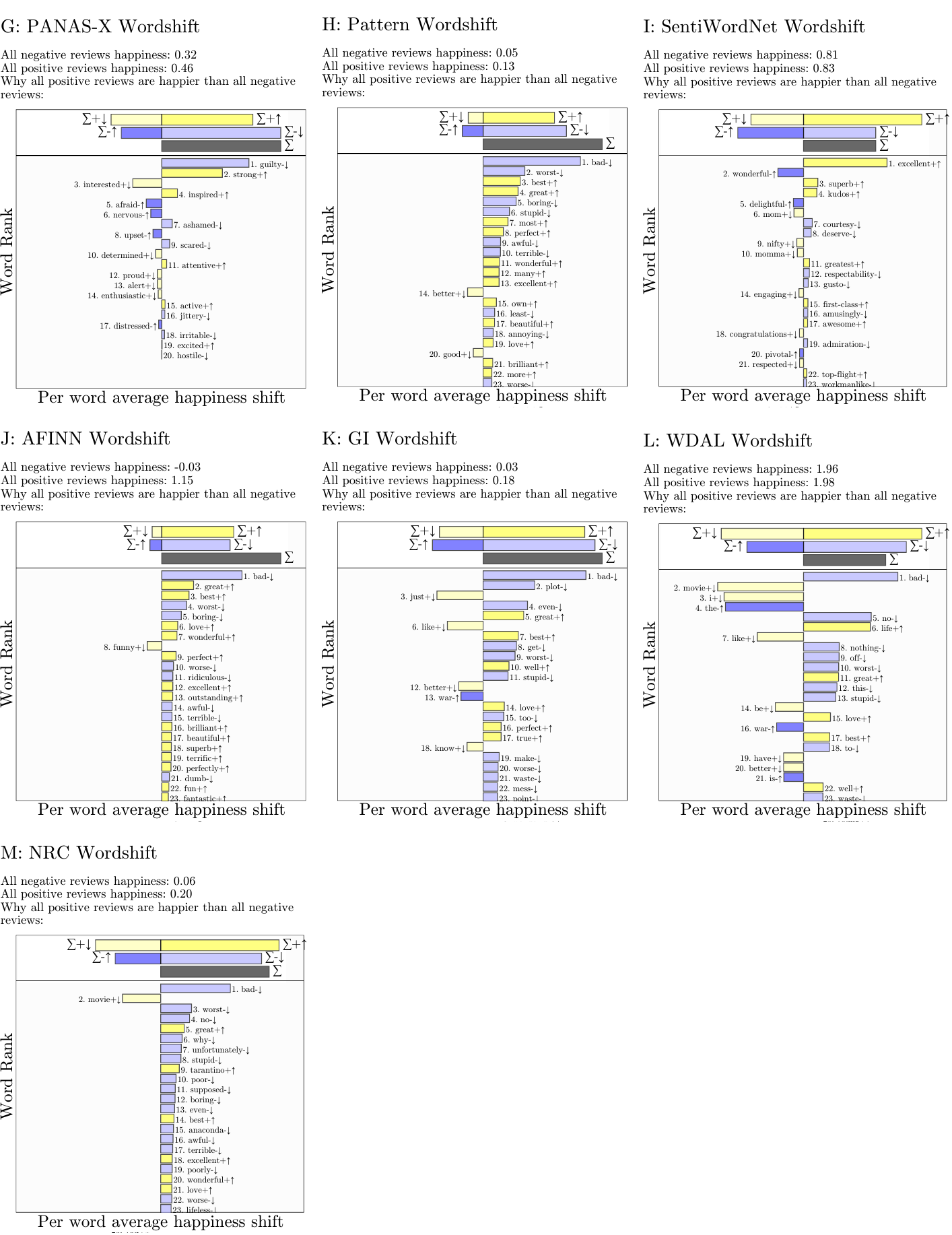}
    \caption[]{
      Word shifts for the movie review corpus, with panel letters continuing from Fig. \ref{fig:moviereviews-shifts}.
      We again see many of the same patterns, and refer the reader to Fig. \ref{fig:moviereviews-shifts} for a more in depth analysis.
  }
  \label{fig:moviereviews-shifts-extras}
\end{figure*}

\clearpage
\pagebreak

\section*{S10 Appendix: Coverage removal and binarization tests of labMT dictionary } \label{supp:labMT-test}

    Here, we perform a detailed analysis of the labMT dictionary to further isolate the effects of dictionary coverage and scoring type.
    This analysis is motivated by ensuring that the our results are not confounded entirely by the quality of the word scores across dictionaries, such that the effect of coverage and scoring type are isolated.
    We focus on the Movie Review corpus for this analysis, and report the accuracy of the labMT dictionary with the aforementioned modifications using the F1 score.

\begin{figure*}[!htb]
  \includegraphics[width=0.48\textwidth]{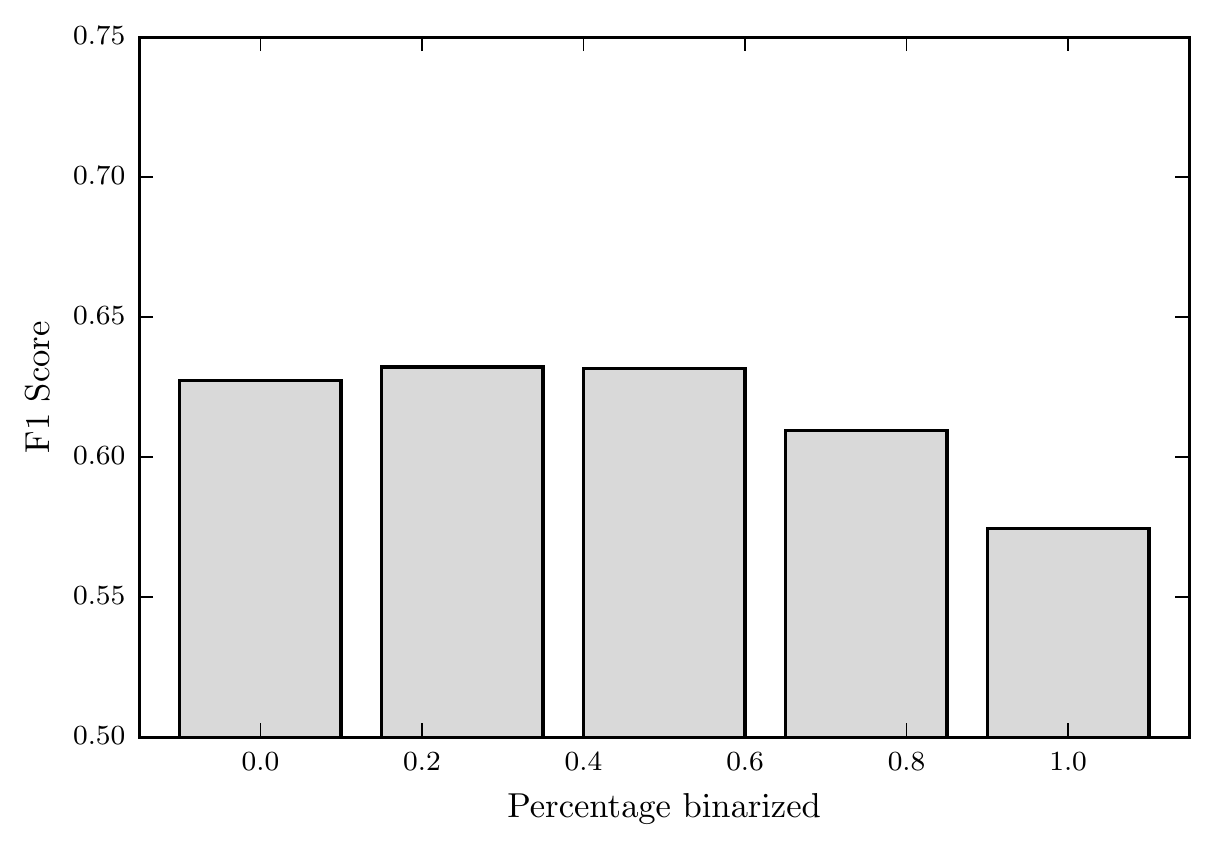}
  \caption[]{
    The direct binarization of the labMT dictionary results in a degradation of performance.
    The binarization is accomplished by linearly reducing the range of scores in the labMT dictionary from a centered -4 $\to$ 4 to the integer scores $-1$ and $1$.
  }
  \label{fig:labMT-binary}
\end{figure*}

\begin{figure*}[!htb]
  \includegraphics[width=0.48\textwidth]{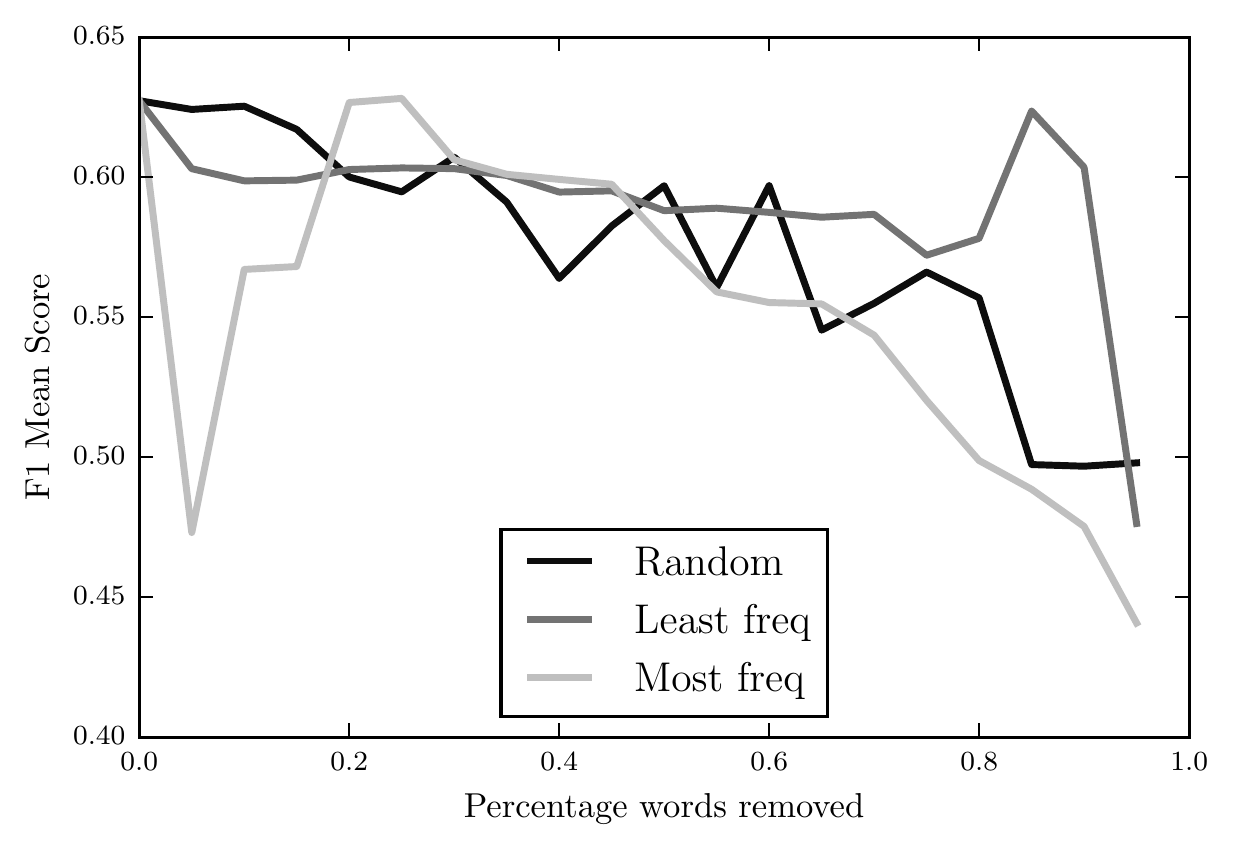}
  \caption[]{
    The resulting F1 score of classification performance for each of three coverage removal strategies.
    These strategies, labeled in the above, are: (1) removing the most frequent words, (2) removing the least frequent words, and (3) removing words randomly (irrespective of their frequency of usage).
  }
  \label{fig:labMT-coverage-removal-result}
\end{figure*}

\begin{figure*}[!htb]
  \includegraphics[width=0.48\textwidth]{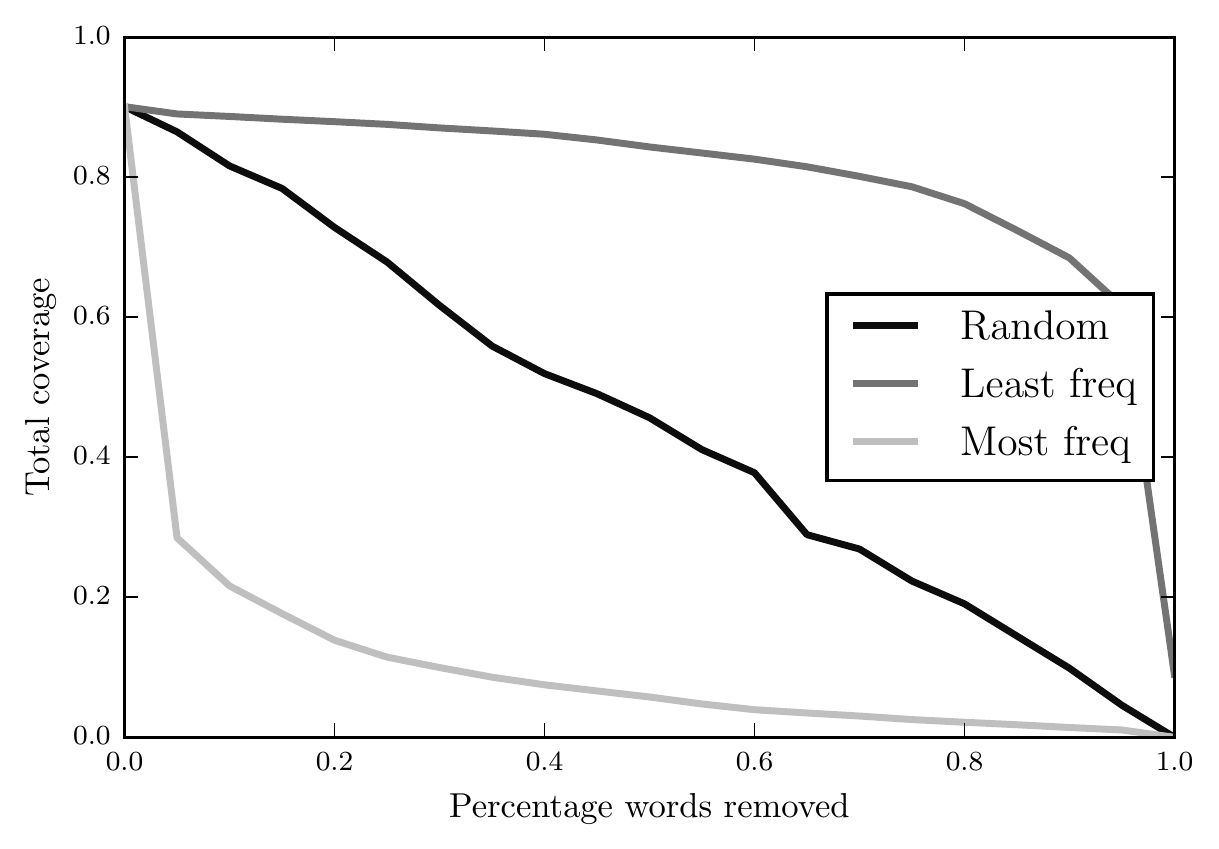}
  \caption[]{
    The resulting coverage for each of three coverage removal strategies.
    Again, these strategies, labeled in the above, are: (1) removing the most frequent words, (2) removing the least frequent words, and (3) removing words randomly (irrespective of their frequency of usage).
  }
  \label{fig:labMT-coverage-removal}
\end{figure*}

    First, we gradually reduce the range of scores in the labMT dictionary from a centered -4 $\to$ 4, down to just the integer scores $-1$ and $1$.
    In Figure \ref{fig:labMT-binary}, the F1 score is show across this gradual, linear change to a binary dictionary.
    We observe that the direct binarization of the labMT dictionary results in a degradation of performance.
    Second, to test the effect of coverage alone, we systematically reduce the coverage of the labMT dictionary and again attempt the binary classification task of identifying Movie Review polarity.
    Three possible strategies to reduce the coverage are (1) removing the most frequent words, (2) removing the least frequent words, and (3) removing words randomly (irrespective of their frequency of usage).
    In Figures \ref{fig:labMT-coverage-removal-result} and \ref{fig:labMT-coverage-removal}, we show the resulting F1 score of classification performance for each of these three strategies and the total coverage from each removal strategy.
    We observe that while certain strategies are more effective at retaining performance, lower coverage scores are all lower despite substantial variation, and the overall pattern for each strategy is a decrease in performance for decreasing coverage.
    In both cases these results are consistent with those seen across dictionaries: integer scores and low coverage strongly reduce the performance of the 2-class movie review classification task, as measured by the F1-score.

\end{document}